%% file: neurips_2022.tex
\documentclass{article}

     \PassOptionsToPackage{numbers, compress}{natbib}

\usepackage[final]{neurips_2022}

\usepackage[utf8]{inputenc} 
\usepackage[T1]{fontenc}    
\usepackage{hyperref}       
\usepackage{url}            
\usepackage{booktabs}       
\usepackage{amsfonts}       
\usepackage{nicefrac}       
\usepackage{microtype}      
\usepackage{xcolor}         
\usepackage{amsmath}
\usepackage{booktabs}
\usepackage{url}
\usepackage{tikz}
\usepackage{caption}
\usepackage{amssymb}
\usepackage{dsfont}
\usepackage{color}
\usepackage{subcaption}
\usepackage{multirow}
\usepackage{float}
\usepackage{mathabx}
\DeclareMathOperator{\argmax}{argmax} 

\title{Value-based CTDE Methods in \\ Symmetric Two-team Markov Game: \\ from Cooperation to Team Competition}

\author{%
  Pascal Leroy \\
  Montefiore Institute, \\
  ULiège \\
  \texttt{pleroy@uliege.be}\\
  \And
  Jonathan Pisane \\
  Thales \\
  \texttt{jonathan.pisane@}\\\texttt{be.thalesgroup.com}
  \And
  Damien Ernst \\
  Montefiore Institute, ULiège \\
  LTCI, Telecom Paris,\\
  Institut Polytechnique de Paris \\
  \texttt{dernst@uliege.be} \\
}

\begin{document}
\usetikzlibrary{positioning, shapes.geometric, arrows.meta, calc}

\maketitle

\begin{abstract}
In this paper, we identify the best learning scenario to train a team of agents to compete against multiple possible strategies of opposing teams.
We evaluate cooperative value-based methods in a mixed cooperative-competitive environment.
We restrict ourselves to the case of a symmetric, partially observable, two-team Markov game.
We selected three training methods based on the centralised training and decentralised execution (CTDE) paradigm: QMIX, MAVEN and QVMix.
For each method, we considered three learning scenarios differentiated by the variety of team policies encountered during training.
For our experiments, we modified the StarCraft Multi-Agent Challenge environment to create competitive environments where both teams could learn and compete simultaneously.
Our results suggest that training against multiple evolving strategies achieves the best results when, for scoring their performances, teams are faced with several strategies.
\end{abstract}


\section{Introduction}
\label{Introduction}
Many applications exist where two teams of multiple agents compete, such as in games (Pommerman \citep{resnick2018pommerman}) or in robotics (RoboCup \citep{kitano1997robocup}).
In certain use-cases, agents are not fully aware of the entire environment, such as in Cyber Physical Production Systems \citep{phan2020learning}, Hide and Seek \citep{baker2019emergent} or Capture the Flag \citep{jaderberg2019human}.
To solve the challenges faced within all these applications, reinforcement learning (RL) is a solution.
RL is a paradigm of machine learning where an agent interacts with an environment by selecting actions based on its observations and receives rewards \citep{sutton2018reinforcement}.
Multi-agent RL (MARL) is an extension of single-agent RL (SARL) where a single agent acts in the environment.
The value-based method is a family of RL methods where an agent learns the highest sum of rewards (value) it can achieve by selecting a given action at a given state.
We distinguish different MARL settings based on the goals of agents. 
They can share the same goal and cooperate, or compete for an opposite goal to the detriment of other agents.
In this paper, we evaluate value-based methods designed for cooperative settings to train agents in a partially observable mixed cooperative-competitive environment where two teams of cooperative agents compete to achieve an opposing goal. Our objective is to identify how to train a team to be resilient to different adversarial strategies.

The cooperative setting can be considered as a decentralised partially observable Markov decision process (Dec-POMDP) \citep{DecPomdp}, which is a framework where agents have partial observations and share the same rewards.
An example is the StarCraft Multi-Agent Challenge (SMAC) \citep{samvelyan2019starcraft}.
The partial observability of agents implies that the control is prevented from being centralised.
This means that it is not possible to control all agents of a team as if they were a single agent.
Therefore, decentralisation is required and each agent must select actions independently based on its own observations.
A first solution is to train each agent of the team independently using SARL methods.
Another possibility is the centralised training with decentralised execution (CTDE), where information about all agents is exploited during training but not during execution.
CTDE methods have shown better results than independent methods in Dec-POMDP.
In this paper, we selected three CTDE value-based methods QMIX \citep{Rashid2018}, MAVEN \citep{Mahajan2019MAVEN:Exploration} and QVMix \citep{leroy2020qvmix}. Other existing methods are discussed in Section \ref{sec:related}.

RL has not been the first choice to solve complex competitive two-player games.
For example, we think of Deep Blue \citep{campbell2002deep}.
However, the improvements in hardware has made possible to apply RL algorithms to play such games with partial observability \citep{silver2018general} or even games with more than two competing agents, such as Quake III Arena in Capture the Flag mode \citep{jaderberg2019human} or StarCraft \citep{vinyals2019grandmaster}.
To train such agents to compete, a paradigm called self-play is used.
Agents play against themselves, allowing them to face evolving strategies and to improve their skills.
In more complex games, a population of learning agents is created by duplicating some of the agents during training to enforce the diversity of opponents.
This allows agents to improve their skills while remaining competent against previously encountered strategies.
In most of the mixed cooperative-competition papers mentioned before, agents are trained with independent SARL methods in self-play or within a population, except for Baker et. al. \cite{baker2019emergent} who chose a CTDE method that exploit the state of the environment during training.

In this paper, we mix cooperative and competitive methods to train teams to compete in a symmetric two-team Markov game where two teams composed of the same agents compete.
Specifically, we study the difference between three learning scenarios: learning against a stationary team policy, against a single evolving team strategy (self-play), and against multiple evolving team strategies within a population of learning teams.
To perform these experiments and to study the performances of each learning scenario, we created a new competitive environment by modifying SMAC which has been designed for cooperation.
We chose symmetrical competition to ensure fair and balanced competition in addition to the possibility of controlling either team with the same agents. 
In two different SMAC environments, we trained teams with three value-based CTDE methods, QMIX, MAVEN and QVMix, each with the three learning scenarios and we then analysed how they perform when faced with multiple opposing strategies.
Our results suggest that when competing with several possible strategies, teams trained in a population achieve the best performance, but a selection process is required to select the best team.
We reached this conclusion irrespective of whether or not the stationary strategy was better than all trained teams.

We define the two-team Markov game and CTDE value-based methods in Section \ref{sec:background}.
The competitive SMAC is described in Section \ref{sec:SMAC}.
In Section \ref{sec:Learning}, we detail the learning scenarios and the performances criteria.
Details of the experiments are provided in Section \ref{sec:experiments} and their results in Section \ref{sec:results}.
The related work is presented in Section \ref{sec:related} followed by the conclusion in Section \ref{sec:conclusion}.

\section{Background}
\label{sec:background}
Our framework is a special case of a Markov game \citep{MarkovGames}.
Specifically, we are in a symmetric, mixed cooperative-competitive, partially observable two-team Markov game defined by a tuple $[\mathcal{S}, O, \mathcal{Z}, \mathcal{U}, n, R, P, \gamma]$, where two teams of $n$ agents each compete by choosing an action at every timestep $t$.
Let $\mathcal{I}=\{1,..,n\}$ and $\mathcal{J}=\{1,-1\}$, the $i^{th}$ agent of the $j^{th}$ team is denoted by $a_{i, j} (i \in \mathcal{I}, j \in \mathcal{J})$ when it is not shortened to $a$.
Its action space is defined by $\mathcal{U}_{a_{i, j}}$.
Since we assume team symmetry, agents $a_{i, 1}$ and $a_{i, -1}$ share the same action space ($\forall i$).
The joint set of $n$ action spaces is denoted by $\mathcal{U}=\bigtimes_{i\in \mathcal{I}} \mathcal{U}_{{a_{i, 1}}} = \bigtimes_{i\in \mathcal{I}} 
\mathcal{U}_{{a_{i, -1}}}$.
The state of the environment at timestep $t$ is $s_t \in \mathcal{S}$ where $\mathcal{S}$ is the set of states, and $o_t^{a} \in \mathcal{Z}$ is the observation of this state perceived by the agent $a$, where $\mathcal{Z}$ is the set of observations. The observation function $O: \mathcal{S} \times I \times J \rightarrow \mathcal{Z}$ determines $o_t^a$.
At each timestep $t$, each agent simultaneously executes an action $u_t^{a} \in \mathcal{U}_a$ such that the state $s_t$ transits to a new state $s_{t+1}$ with a probability defined by the transition function $P(s_{t+1}|s_t, \boldsymbol{u_t}): \mathcal{S}^2 \times \mathcal{U}^2 \rightarrow \mathbb{R^+}$ where the joint action $\boldsymbol{u_t} = \{\boldsymbol{u^1_t}, \boldsymbol{u^{-1}_t}\} = \bigcup_{i \in \mathcal{I}, j \in \mathcal{J}} u_t^{a_{i, j}}$. 
After the joint action is executed, common team rewards $\{r_t^1, r_t^{-1}\} = R(s_{t+1}, s_t, \boldsymbol {u_t}): \mathcal{S}^2 \times \mathcal{U}^2 \rightarrow \mathbb{R}^2$ are assigned to agents.
An agent $a$ chooses its action based on its current observation $o_t^{a} \in \mathcal{Z}$ and its history $\tau_t^{a} \in (\mathcal{Z} \times \mathcal{U}_a)^{t-1}$.
Its policy is the function $\pi^{a}(u_t^{a}|\tau_t^{a},o_t^{a}): (\mathcal{Z} \times \mathcal{U}_a)^t \rightarrow \mathbb{R^+}$ that maps, from its history and the current observation, the probability of taking action $u_t^{a}$.
The team policy is denoted by $\boldsymbol {\pi_j}=\bigcup_{i \in \mathcal{I}} \pi^{a_{i, j}}$.

During an episode, the cumulative discounted reward obtained from timestep $t$ over the next $T$ timesteps by team $j$ is defined by $R_{t}^j = \sum_{k=0}^{T-1} \gamma^k r^j_{t+k}$ where $\gamma \in [0, 1)$ is the discount factor.
The goal of each agent is to find the optimal team policy that maximises the expected cumulative discounted reward: $\boldsymbol {\pi_j^{*}} =\argmax_{\boldsymbol {\pi_j}} \mathds{E}[R^j_0|\boldsymbol{\pi_{j}}, \boldsymbol{\pi_{-j}}]$ that depends on the other team policy $ \boldsymbol {\pi_{-j}}$.
To compute the optimal policy, in value-based methods, we rely on the value function $V^{\boldsymbol {\pi}, j}(s)=\mathds{E}[R^j_t | s_t = s, \boldsymbol {\pi}]$, called the $V$ function.
We also define the state-joint-action value function $Q^{\boldsymbol {\pi}, j}(s,\boldsymbol{u})=\mathds{E}[R^j_t | s_t=s, \boldsymbol{u^j_t}=\boldsymbol{u}, \boldsymbol {\pi}]$, called the $Q$ function.
The corresponding individual state-action value function of an agent policy is defined by $Q^{\boldsymbol{\pi}, a_{i, j}}(s,u)=\mathds{E}[R^j_t | s_t=s, u^{a_{i, j}}_t=u, \boldsymbol{\pi}]$ and denoted as $Q_a$ for the sake of conciseness. Since agents share the same reward in the same team, $Q^{\boldsymbol{\pi}, a_{i, j}}(s,u)=Q^{\boldsymbol{\pi}, j}(s,\boldsymbol{u})$.
Note that the symmetry allows one to control both teams with the same policy.

It is possible to obtain the Dec-POMDP \citep{DecPomdp} definition from that of the two-team Markov game by constraining one team to have a stationary policy.
The other team would then be the only team learning and the two-team Markov game can be considered as a Dec-POMDP.
Its formal definition is easily derived from that of the two-team Markov game by eliminating team considerations and ignoring the $j$.
In our experiments, teams are trained with value-based methods designed for Dec-POMDP.
As stated in the introduction, it is not possible to learn only the state-joint-action value function $Q(s,\boldsymbol{u})$ to centrally control the agents because agents must select their action based only on $(o, \tau)$ and do not perceive $s$.
A solution is known as independent Q learning (IQL) \citep{Tan1993} and consists in training agents to learn $Q_a$ independently, ignoring the presence of other learning agents.
Another solution is to exploit more information during training but only the available $(o, \tau)$ during execution.
It is a paradigm called centralised training with decentralised execution (CTDE).
The algorithms tested in this paper, QMIX \citep{Rashid2018}, MAVEN \citep{Mahajan2019MAVEN:Exploration} and QVMix \citep{leroy2020qvmix} exploit this paradigm.
All three methods learn $Q(s_t,\boldsymbol{u_t})$ as a factorisation of $s_t$ and $Q_a$ $\forall a$ during training.
The $Q_a$ are no longer $Q$ functions but are utility functions used to select actions which are retained for the execution.
There exists a condition to factorise $Q(s_t,\boldsymbol{u_t})$, called the Individual-Global-Max condition (IGM), that requires that $\argmax_{\mathbf{u_t}} Q(s_t, \mathbf{u_t}) =\bigcup_{a}\argmax_{u_t^{a}} Q_{a}$ \citep{Son2019QTRAN:Learning}.
In QMIX, $Q_a$ factorise $Q(s_t,\boldsymbol{u_t})$ with a hypernetwork.
The weights of each $Q_a$ in the factorisation are computed based on $s_t$ and are constrained to be positive to ensure IGM.
MAVEN follows the same principle of QMIX with an additional hierarchical policy network that modifies each $Q_a$ in order to influence agents' behaviour and to improve their exploration capacity.
QVMix is an extension of the Deep Quality-Value (DQV) family of algorithms \citep{sabatelli2018deepQV,sabatelli2020deep} applied to QMIX where agents learn both the $Q$ and the $V$ functions.
Those value-based methods which were initially developed for solving Dec-POMDP are thoroughly detailed in Appendix \ref{app:vbmethod}.
We chose QMIX because of its popularity, MAVEN because of its exploration technique which outperforms QMIX in complex scenarios, and QVMix because it has proven to be competitive with these previously mentioned algorithms.
Other methods have been developed as described in Section \ref{sec:related}.

\section{Competitive StarCraft Multi-agent challenge}
\label{sec:SMAC}

To perform our experiments, we created a new environment by modifying the StarCraft Multi-Agent Challenge (SMAC) \citep{samvelyan2019starcraft} to create a competitive environment. 
SMAC has been developed to train a team to cooperate in a competitive environment but only against the built-in StarCraft AI which has a stationary strategy. 
We adapted\footnote{\label{foot_note_code}\url{github.com/PaLeroy/competSmac} - \url{github.com/PaLeroy/competPymarl}} both SMAC and the associated learning framework PyMARL \citep{samvelyan2019starcraft}.
It is now possible to control both opposing teams in SMAC, as well as train them simultaneously.
In competitive SMAC, the goal remains unchanged, and it is to defeat the opposing team, by inflicting sufficient damage to reduce the hit points of all opponents to zero.
The winning team is the one that ends up with the highest sum of remaining hit points at the end.
If both teams end up with the same total of hit points, it is a draw.
In SMAC, a map is the name given to the different scenarios, and from an RL perspective, a different map is a different environment.
For our experiments, we have chosen the $3m$ and $3s5z$ maps.
In the $3m$ map, two teams of three marines compete for a maximum $100$ timesteps.
In the $3s5z$ map, two teams of three stalkers and five zealots compete for a maximum $150$ timesteps.
More details about these two maps and competitive SMAC can be found in Appendix \ref{app:smac}.

\section{Learning scenarios and performances criteria}
\label{sec:Learning}
Our goal is to train a team to be resilient to different adversarial strategies.
We test three different learning scenarios differentiated by the diversity of opponents' strategies encountered during training.
Thanks to the environment symmetry, the trained teams can act as any of the two teams in our environments.
In the first learning scenario, the team is trained against a stationary strategy, which we refer to as a heuristic.
As introduced in Section \ref{sec:background}, such configuration is a Dec-POMDP, framework for which CTDE methods are designed.
The second scenario is called self-play where the team is trained by playing against itself and thus facing a strategy that continuously improves at the same learning speed.
This scenario has proven its value in competitive MARL \citep{baker2019emergent}.
In the third scenario, a population composed of several teams is trained with the same method.
Teams either play against themselves or against other learning teams.
Self-play is the special case of a training population composed of a single team.
This learning scenario has proven itself when the number of winning strategies is large, such as in StarCraft \citep{vinyals2019grandmaster}.
The latter trained a growing population of agents but in this paper, the size of the training population is fixed to five to reduce computational complexity.

After training, we evaluate teams with the Elo rating system \citep{elo1978rating}.
Its purpose is to assign each player of a population with a rating.
From these ratings, one can compute the probability that a player will win when facing another one.
Individuals' scores are updated based on the outcome of the episode  and their current Elo scores.
The Elo rating system is formally described in Appendix \ref{app:elo}.

We form test populations to compute Elo scores in different configurations.
To evaluate only the learning scenarios, we form one test population for each CTDE method with teams trained with all learning scenarios.
To evaluate the performances of the heuristic, we add it in the previously defined test populations to form new ones.
To find the best learning scenario/training method pair, we group all trained teams in two test populations, with and without the heuristic.
To evaluate training efficiency and heuristic performances, each trained team is tested along with training against the heuristic and against teams trained with other learning scenarios but using the same CTDE method.

\section{Experiments}
\label{sec:experiments}

For our experiments, teams were trained with three learning scenarios (Sec. \ref{sec:Learning}) and with three CTDE methods, QMIX, MAVEN and QVMix (Sec. \ref{sec:background}), in two different environments (Sec. \ref{sec:SMAC}).
We conducted the same experimental process for both environments.
For each method, each learning scenario was executed ten times and stopped after each team had exploited $10^7$ samples.
For each method, this resulted in ten teams trained against the heuristic requiring $10^7$ environment timesteps, ten teams trained in self-play requiring $5\times10^6$ environment timesteps and ten training populations of five teams leading to $50$ teams trained within populations requiring at least $5\times10^7$ environment timesteps.
Therefore, in total, $210$ teams of nine types were trained in both environments.

To train teams against the heuristic, it was ensured that they play the same number of episodes as team $j=1$ and team $j=-1$.
Concerning the training within a population, each team had an equal chance of playing as team $j=1$ or team $j=-1$, meaning an equal chance to play against any team in the population, including itself.
For all learning scenarios and methods, network architectures and parameters are the same to ensure fair comparison.
Learning parameters were determined by default configurations provided by their different authors \citep{Rashid2018,Mahajan2019MAVEN:Exploration,leroy2020qvmix}.
More details about the training parameters is provided in Appendix \ref{app:train_param}.
The main differences with the original are that the epsilon anneal time is set to $2$ million instead of $0.5$ million and that the networks are updated every eight episodes in $3m$ and every episode in $3s5z$.
As in the literature, individual networks of each team share the same parameters to improve learning speed. We present training times in Appendix \ref{app:train_time}.

In this paper, the heuristic is based on two rules.
It moves toward the starting point of the opponent's team until it reaches the opposite side of the map and stops.
If there are enemies within its shooting range, it attacks the nearest one.
This heuristic is slightly different from StarCraft's built-in AI originally used to train teams to cooperate in SMAC.
The built-in AI also moves toward the other side but selects targets based on a priority score.
It will choose to attack the closest unit with the highest priority and this unit will remain the target until its priority drops or it can no longer be attacked.
A unit's priority score is based on its type and current action.
For example, if two of the same units attack and the targeted unit stops attacking, its priority score will drop and the built-in AI will select the other unit to attack.
This is the main difference with our heuristic which will attack the nearest unit regardless of its action and its priority.
Results show that our heuristic is harder to beat than the one provided in SMAC.
Finally, while both maps are denoted as easy in \citep{samvelyan2019starcraft}, the task of learning everything from scratch is not.
When learning against the heuristic, teams do not need to learn how to find their opponents, because they automatically move towards them.
When they learn in self-play or within a population, they first need to learn where to find opponents before they learn to face them.
This increase in learning complexity compared to their cooperative version led us to this choice of maps, motivated by a compromise between computational complexity and learning complexity.

As described in Section \ref{sec:Learning}, we form test populations to evaluate teams with the Elo rating system.
For both environments, we have three test populations of 70 teams trained with the same method and the three learning scenarios.
We also have one test population with all the 210 trained teams.
We created four additional test populations from the four previously defined by adding the heuristic to them.
In practice, to compute the Elo scores in these 16 test populations, each team plays $20$ games against all the other teams in a randomised order.
Every team starts with an Elo score of $1000$ and we set the maximum Elo score update to $10$, which is sufficiently small for our population sizes.
In Section \ref{sec:results}, we analyse the distribution of Elo scores after every team had finished its testing games.

During training, team neural network parameters are recorded every $2\times10^5$ timesteps until the $10$ millionth played timestep.
The test populations are composed of the agents whose network parameters correspond to those recorded at the 10 millionth timestep.
The other saved networks allow one to evaluate teams' performances during training.
We evaluate teams trained with a method and a learning scenario against the heuristic and against all teams trained with the same method but a different learning scenario.
We analysed the win rates along training timesteps of these different matchups when teams play $24$ games against each other.
For example, and for the same method, each ten teams trained in self-play played $24$ games against all the $50$ teams trained within a population and against the $10$ teams trained against the heuristic.

\section{Performances of learning scenarios}
\label{sec:results}

We present the Elo scores of teams with box plots in Figure \ref{fig:elo_method} when the test population contains teams trained with the same method.
We first focus on performances when the heuristic is not in the test population (Fig. \ref{subfig:elo_no_h_3m}, \ref{subfig:elo_no_h_3s5z}).
The first observation is that the best teams are the ones trained within a population, except with MAVEN in $3s5z$ (Fig. \ref{subfig:3s5z_elo_no_h_methodMAVEN}) for which the differences between learning scenarios are smaller.
We discuss these differences later.
The second observation is that the population scenario has the highest variance.
To understand this, we also plot the box plots corresponding to the ten teams that achieved the highest Elo score of each training population, denoted by \textbf{BP}.
These box plots confirm that there is a difference between teams of the same training population and a selection must be performed to find the best one, so as to optimise the performances of this learning scenario.
Training against the heuristic is the worst scenario, arguably because agents do not generalise to other strategy than the heuristic.
However, the heuristic is not in these test populations and its impact is the concern of later analysis.
The performances of teams trained in self-play lies in between the two other learning scenarios.
While some teams achieve Elo scores close to the best teams of each training population, the scores of others are lower than the lowest scores of teams trained within population.

\begin{figure*}[h]
\input{fig_elo} 
\end{figure*} 

The same experiment is performed with the addition of the heuristic in the three test populations and corresponding box plots are presented in Figures \ref{subfig:elo_h_3m} and \ref{subfig:elo_h_3s5z}.
The heuristic scores are different depending on the map.
In the $3m$ map, most of the teams achieve a higher Elo score than the heuristic, whereas in $3s5z$, the heuristic dominates all teams.
In the $3s5z$ map and for all three learning scenarios, Elo scores slightly decreased with the addition of the heuristic.
The conclusion is straightforward and the heuristic is better than all teams.
However, one should note that the score ordering between the teams remained the same between Figure \ref{subfig:elo_no_h_3s5z} and \ref{subfig:elo_h_3s5z}.
In $3m$, one can see that the Elo score of teams trained against the heuristic (\textbf{H}) is higher in Fig. \ref{subfig:elo_h_3m} than the ones in Figure \ref{subfig:elo_no_h_3m}, as a direct consequence of the introduction into the test populations of a team against which they win.
When compared with the previous test populations (Fig. \ref{subfig:elo_no_h_3m}), teams trained with QVMix achieved higher Elo scores than without the heuristic in the test population (Fig. \ref{subfig:elo_h_methodQVMIX}), while when trained with MAVEN and QMIX, they achieved lower Elo scores (Fig. \ref{subfig:elo_h_methodMAVEN}, \ref{subfig:elo_h_methodQMIX}).
In all cases, the higher values of box plots are not significantly different but lower values are, meaning that some teams performed poorly against the heuristic.
The conclusion is that teams trained within a population remain the most successful in most of our experiments, no matter if the heuristic is the best or almost the worst team.

\begin{figure*}
    \input{fig_training}
\end{figure*}

In Figure \ref{subfig:3m_vs_h}, we present the evolution of win rates against the heuristic along training timesteps by each team in the $3m$ map.
For all methods, teams trained against the heuristic are the best against it on average.
This explains why their Elo scores improve when the heuristic is included in the test population.
This is also the case for QVMix teams trained in self-play and within a population that performed better and learned faster than teams trained with MAVEN and QMIX with these two learning scenarios.
However, in the $3s5z$ map, it can be seen in Figure \ref{subfig:3s5z_vsh} that the win rates against the heuristic are very low, not to say equal to zero.
Only the win rates of teams trained against the heuristic, especially with QVMix and QMIX, increase at the end of the training but with a high variance in comparison to the $3m$ map (Fig. \ref{subfig:3m_vs_h}).
For QVMix, the win rates of teams trained with a population also increase at the end of the training phase.
The time we have budgeted for training in the $3s5z$ map may be insufficient to achieve a high win rate.
However, we find this beneficial because it shows that, even when the heuristic is better than all teams, training against it, with the same training timesteps allowance, is not the best learning scenario when teams have to be good against several strategies.
This also shows that our heuristic is harder to defeat with respect to the results of \citep{Rashid2018,Mahajan2019MAVEN:Exploration,leroy2020qvmix} where better results are observed in these maps with the former SMAC heuristic.

In the $3m$ map, the standard deviation of green and blue win rates, representing population and self-play learning scenarios respectively, is high and confirms the results of Elo score box plots that there are performance gaps between teams trained within the same learning scenario.
Observations also suggest that teams trained in self-play would achieve the same win rates as teams trained within a population if they were trained for longer, suggesting a difference of training sample efficiency.
As teams first need to learn to cross the map to meet opponents and to learn to fight them, this difference is because training within a population, against several strategies, increases the probability of creating episodes where two opposing agents meet and fight at the beginning of training.

Win rates from the confrontations between trained teams are presented in Figures \ref{subfig:3m_duo} and \ref{subfig:3s5z_duo}.
The draw rates can be obtained by subtracting from $1$ the sum of these two curves.
This confirms the previous results with box plots that training against the heuristic is the worst scenario, as an average win rate above $60\%$ is achieved against them as shown in the black and red curves.
Green and blue curves enable one to analyse self-play against population-learning scenarios.
At the beginning of the training phase, teams trained within a population are better than self-play ones in the $3m$ map.
This high win rate decreases with training in favour of the win rate of teams trained in self-play, and for QVMix and QMIX, until it becomes higher than the latter.
Again, this suggests a difference in training sample efficiency.
Moreover, although the training sample efficiency is lower for teams trained in self-play, the number of environment timesteps required to train them is five-times lower in our setting.
However, it is, on average, that the self-play teams become better.
In the $3s5z$ map, this overlap phenomenon does not occur and the average win rates fluctuate around $50\%$ with the green curves remaining just above the blue ones at the end.
The proximity of performances between teams trained in self-play and within a population is arguably due to the environment and the $3m$ map which does not offer the possibility of winning with very different strategies.
As the $3s5z$ map appears to be more complex, with the lack of performances against the heuristic as evidence, teams trained within a population remain better on average.

On average, the results of the teams trained within a population with MAVEN in the 3s5z map differ from the other experiments. 
Some results are slightly worse in regard to those of the other experiments.
The reason for this is not clear, as we executed the experiments several times, but this does not affect the conclusions of our experiments that remain clear.
Finally, it would be necessary to repeat these experiments more times or to analyse the behaviour of the agents in depth to find the problem, which is beyond the scope of this paper. 

\begin{figure*}[ht]
\begin{subfigure}{\textwidth}
\centering
\includegraphics[width=.95\textwidth]{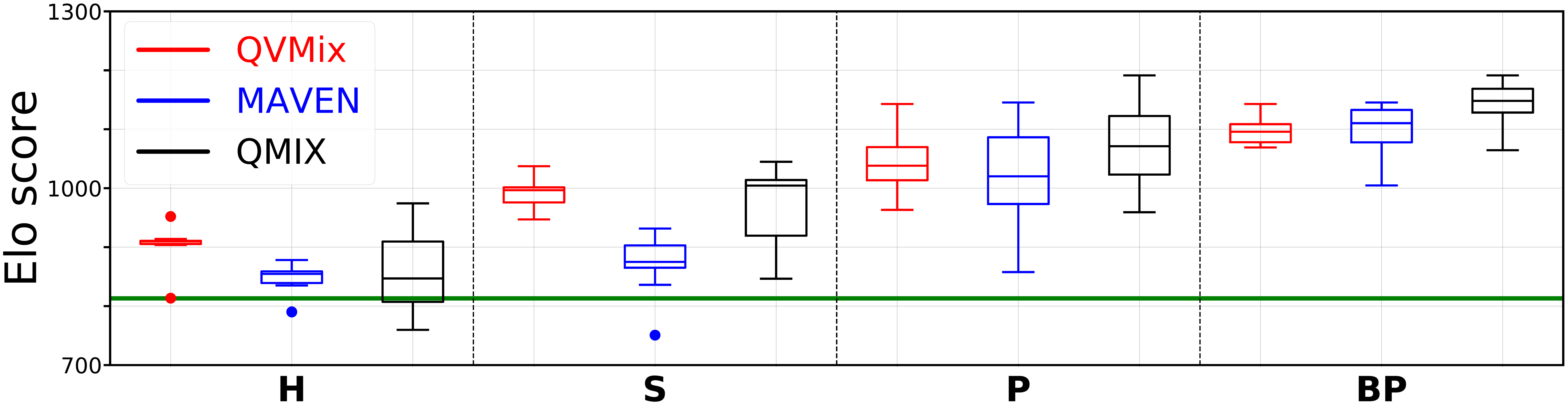}
\caption{$3m$}
\label{subfig:3m_all_h}
\end{subfigure}
\begin{subfigure}{\textwidth}
\centering
\includegraphics[width=.95\textwidth]{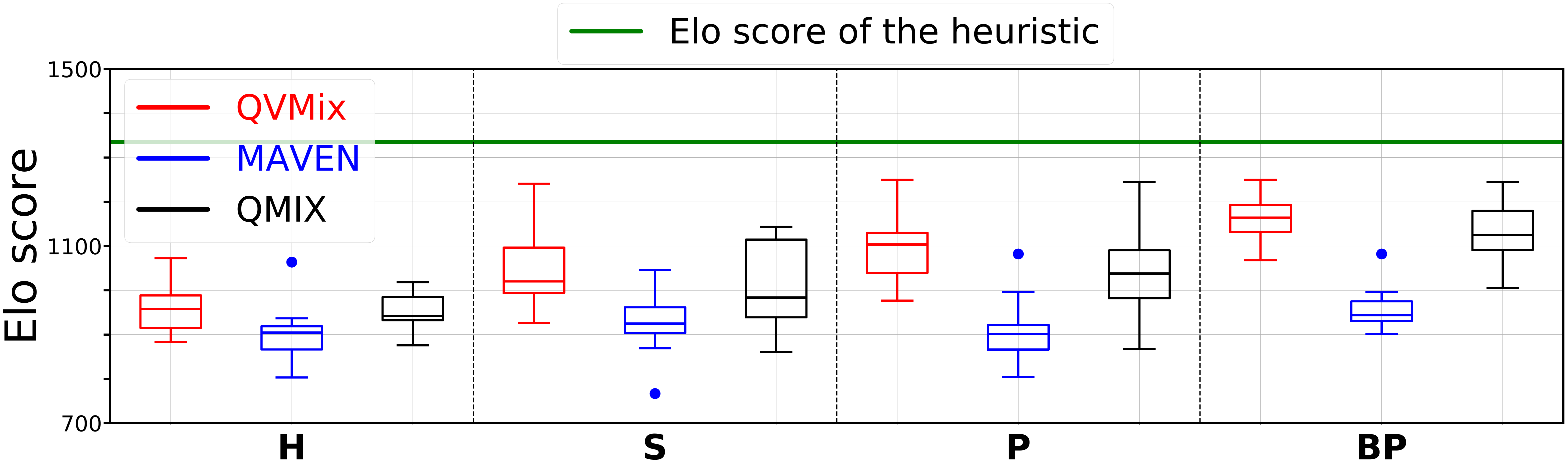}
\caption{$3s5z$}
\label{subfig:3s5z_all_h}
\end{subfigure}
\caption{
Elo score box plots of two test populations, in $3m$ at the top and in $3s5z$ at the bottom, composed of the heuristic and teams trained using three methods and three learning scenarios.
The training method is either QVMix (red), MAVEN (blue) or QMIX (black).
Box plots represent the distribution of the ELO scores of teams trained either against the heuristic (\textbf{H}), in self-play (\textbf{S}), within a population (\textbf{P}) or the best of each population (\textbf{BP}).
Box plots present the median, the first quantile ($Q1$) and the third quantile ($Q3$). The reach of whiskers is defined by $1.7*(Q3-Q1)$.
}
\label{fig:all}
\end{figure*}

In Figure \ref{fig:all}, we present the box plots of Elo scores obtained by all trained teams and the heuristic in a single test population for both maps.
We observe that the learning scenario ranking remains the same as in other experiments.
In the $3m$ map, QMIX achieves the highest Elo scores while the lowest MAVEN Elo scores are worse than the ones of QMIX and QVMix when teams are trained in self-play or within a population.
QVMix produces results with a lower variance than QMIX.
In the $3s5z$ map, QVMix is the one achieving the highest Elo scores and it is also MAVEN that achieves the lowest ones.
We conducted the same experiment without the heuristic, which led to the same conclusion (see Figure \ref{fig:all_no_h} in Appendix \ref{app:all_no_h}).
Despite its mechanism of exploration, MAVEN is not able to outperform QMIX and QVMix.
This is also the case in \citep{Mahajan2019MAVEN:Exploration} and \citep{leroy2020qvmix} where they show that MAVEN outperforms QMIX but not QVMix in more complex Dec-POMDP environments.

\section{Related work}
\label{sec:related}
Close to our work, Phan et.al. \citep{phan2020learning} proposed the antagonist-ratio training scheme (ARTS) to train a team to be resilient to failures.
To improve resilience of cooperative methods, they simulate failures as a mixed cooperative-competitive setting where the faulty agents are adversarial agents of another team.
As in our work, ARTS combines CTDE methods, such as QMIX, and population based training.

In this paper, we consider three CTDE value-based methods QMIX \citep{Rashid2018}, MAVEN \citep{Mahajan2019MAVEN:Exploration} and QVMix \citep{leroy2020qvmix} designed for Dec-POMDP.
They rely on the same factorisation of $Q(s, \mathbf{u})$ with a constrained hypernetwork.
This constraint implies representation limits.
Other research papers have sought to change this factorisation and achieved better performances than QMIX.
To cite some, QTRAN \citep{Son2019QTRAN:Learning} factorises it by additive decomposition and QPLEX \citep{wang2021qplex} uses a duplex dueling network architecture, taking advantage of the dueling architecture \citep{wang2016dueling}, to learn both $Q(s, \mathbf{u})$ and $Q_a$.
In parallel, others have succeeded to remove the factorisation, such as LAN \citep{avalos2021local} which also take advantage of the dueling architecture \citep{wang2016dueling} to learn a centralised value function with independent advantage functions.
Policy-based methods is another family of RL methods that have inspired many CTDE algorithms, such as COMA \citep{Foerster2017}, MADDPG \citep{lowe2017multi}, LIIR \citep{Du2019LIIRLearning}, FACMAC \citep{peng2021facmac} and HATPRO, HAPPO \citep{kuba2021trust}.

As introduced, self-play or population based training for competitive settings have already been explored in many works \citep{jaderberg2019human, vinyals2019grandmaster, baker2019emergent}.
More advanced techniques have been widely studied in the literature such as fictitious play \cite{brown1951iterative} or best response dynamic \citep{baudin2022fictitious}.
In short, it consists in choosing the policy which is the best response against the average strategy of an opponent.
In self-play, we mention fictitious play to play Poker \citep{pmlr-v37-heinrich15} and Deep Nash to play Stratego \citep{DM_stratego}.
In population based training, we mention policy-space response oracles (PSRO) \citep{NIPS2017_3323fe11, Muller2020A}.
These methods are referred to as opponent modelling.
Extensions of these works consider to additionally learn how the opponent will update its strategy \citep{he2016opponent, foerster2017learning}.
Recently, Tian et. al. \citep{tian2022multi} introduced a time dynamical opponent model to encode opponent policies in a CTDE framework for mixed cooperative-competitive environment.

\section{Conclusion and future works}
\label{sec:conclusion}
In this paper, we evaluated learning scenarios to train teams to face multiple strategies in a symmetric two-team Markov game.
Teams are trained with three CTDE value-based methods: QMIX, MAVEN and QVMix and with three learning scenarios differentiated by the variety of strategies that teams will encounter during their training.
Specifically, they are trained by playing against a stationary strategy, against themselves, or within a population of teams trained using the same method.
To perform our experiments, we modified the cooperative environment SMAC to allow teams to compete and train at the same time, and trained teams in two different SMAC environments.
These nine types of trained teams are evaluated at the end of their training with the Elo rating system.
Different groups are formed to identify which learning scenario is the best and which learning scenario/training method pair is the best.
We also analysed the win rates of several matchups during training to support the results provided by the Elo scores.
Our results showed that training teams within a population of learning teams is the best learning scenario, when each team plays the same number of timesteps for training purposes.
We reached this conclusion irrespective of whether or not the stationary strategy was better than all trained teams.
Finally, a selection procedure is required because teams from the same training population do not perform equally.

This work is a first investigation of two-team competition with CTDE methods, and we hereafter suggest several future research directions.
First, we suggest to perform the same experiments on more complex environments and to tackle the challenges of an asymmetric two-team Markov games.
In this paper, we selected value-based methods because of their performances in SMAC \citep{samvelyan2019starcraft} at the time of our experiments.
Recent CTDE methods overcome these performances and may lead to interesting new results.
Another research direction would be to study how diversity in the training population impacts the performances. 
Diversity could be increased by adding the heuristic in the population, confronting agents against older learned policies, varying the size of the population or training teams with different methods in the same population.
In addition, as described in Section \ref{sec:related}, methods that compute best responses or model the opponent are widely studied in the literature and would be interesting to test in such environment.
We also propose performing behavioural and policy analyses so as to better understand why some teams achieve a better Elo score and how different are strategies in a single training population.

\begin{ack}
Computational resources have been provided by the Consortium des Équipements de Calcul Intensif (CÉCI), funded by the Fonds de la Recherche Scientifique de Belgique (F.R.S.-FNRS) under Grant No. 2.5020.11 and by the Walloon Region. 
We also acknowledge the financial support of the Walloon Region in the context of IRIS, a MecaTech Cluster project.
\end{ack}

\bibliographystyle{plain}
\bibliography{biblio}

\newpage

\appendix

\section{Value-based methods in a cooperative setting}
\label{app:vbmethod}
Value-based methods consist of learning a Q function.
Formally, in SARL, this corresponds to learning $Q^{\pi^*}(s,u)= \max_{\pi}\:Q^{\pi}(s,u)$  $\forall (s,u) \in \mathcal{S} \times \mathcal{U}$.
The optimal policy is a greedy selection: $u^*=\argmax_u Q^{\pi^*}(s, u)$.
Q-learning \citep{watkins1992q} showed that it is possible, when interacting with MDPs, to learn the exact $Q$ functions using simple update rules depending only on the information collected by the agent.
However, when $\mathcal{S} \times \mathcal{U}$ is very large or continuous, these methods can become intractable and so function approximators are used to model the $Q$ function.
A neural network, parametrised by $\theta$, can be used to learn a $Q$ function approximation by minimising the objective defined in Equation \ref{eq:DQN_loss} where $B$ is the experience replay and $\theta'$ parametrises a target network.

\begin{equation}
    \mathcal{L}(\theta) =  \mathds{E}_{\langle s_{t},u_{t},r_{t},s_{t+1}\rangle \sim B}
    \bigg[  
    \big(r_{t} + \gamma \max_{u \in \mathcal{U}} Q(s_{t+1}, u; \theta') 
     - Q(s_{t}, u_{t}; \theta)\big)^{2}
    \bigg] 
    \label{eq:DQN_loss}
\end{equation}

The experience replay stores transitions $\langle s_{t},u_{t},r_{t},s_{t+1}\rangle$ and the target network is a copy of $\theta$ that is periodically updated. 
In DQN \citep{Mnih2015}, $\theta$ parametrises a convolutional network while in DRQN \citep{Hausknecht2015DeepMDPs}, $\theta$ parametrises a recurrent neural network (RNN), which has been shown to achieve better performances in partially observable environments.
When $\theta$ is a recurrent neural network, $B$ stores sequences of contiguous transition as the update of recurrent neural networks is performed on sequences.

When considering value-based methods in a Dec-POMDP, one possible method to consider is Independent Q-Learning (IQL) \citep{Tan1993}.
With IQL, agents independently learn their $Q$ function, as in SARL, without considering the existence of other learning agents in their environment.
One problem with IQL is that agents must select actions which maximise $Q(s_t, \mathbf{u_t})$ while ignoring, at any time, actions taken by other agents.

This is where CTDE becomes useful. It is possible to approximate $Q(s_t, \mathbf{u_t})$ as a factorisation of individual $Q_a$ functions during training such that $\mathbf{u_t}$ maximises both the joint and the individual $Q_a$ functions.
To ensure this, individual $Q_a$ functions must satisfy the Individual-Global-Max condition (IGM)\citep{Son2019QTRAN:Learning} presented in Equation \ref{eq:igm}.

\begin{equation}
    \argmax_{\mathbf{u_t}} Q(s_t, \mathbf{u_t}) =\bigcup_{a}\argmax_{u_t^{a}} Q_{a}(s_t, u_t^{a})
    \label{eq:igm}
\end{equation}
    
\subsection{QMIX} 
QMIX \citep{Rashid2018} is a CTDE method where the factorisation of $Q(s_t, \mathbf{u_t})$, denoted as $Q_{mix}(s_t, \mathbf{u_t})$, is performed as a function of the individual $Q_a$ functions and the state during training. It is defined in Equation \ref{eq:qmixappendix}.

\begin{equation}
     Q_{mix}=\text{Mixer} \left(Q_{a_1}(s_t, u_t^{a_1}) ,..,Q_{a_n}(s_t, u_t^{a_n}), s_t\right)
     \label{eq:qmixappendix}
\end{equation}

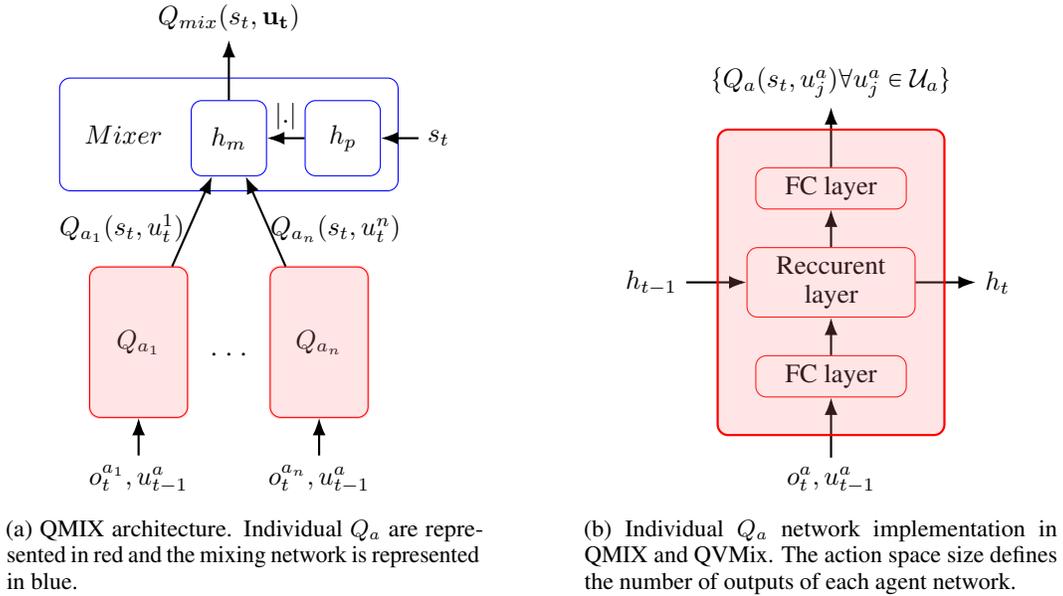
\begin{figure}[ht]
\begin{subfigure}[t]{.45\textwidth}
    \centering
    \input{tikz/qmix}
    \caption{QMIX architecture. Individual $Q_a$ are represented in red and the mixing network is represented in blue.}
    \label{fig:qmix}
    \end{subfigure}%
    \hfill
    \begin{subfigure}[t]{.45\textwidth}
        \centering
    \input{tikz/indivQ}
    \caption{Individual $Q_a$ network implementation in QMIX and QVMix. The action space size defines the number of outputs of each agent network.}
    \label{fig:indivQ}
    \end{subfigure}%
    
\caption{Details of the QMIX and QVMix architecture.}
\end{figure}

The mixer satisfies IGM by enforcing $\frac{\partial Q_{mix}(s_t, \mathbf{u_t})}{\partial Q_{a}(s_t, u_t^{a})} \geq 0 \text{ } \forall a \in \{a_1,..,a_n\}$ by constraining a hypernetwork \citep{Ha2016HyperNetworks} to produce positive weights in order to factorise $Q(s_t, \mathbf{u_t})$.
Formally, this is defined by a hypernetwork $h_p(.): \mathcal{S} \rightarrow \mathbb{R}^{|\phi|+}$ which takes the state $s_t$ as input and computes the strictly positive parameters\footnote{To be exact, the offsets defined by $h_p()$ are not constrained to be positive, only the weights.} $\phi$ of a main network $h_m(.)$.
This main network takes as input all individual $Q_a$ to compute $Q_{mix}$ with the positive weights and the offsets defined by $\phi$.
Together, $h_p(.)$ and $h_m(.)$ defines the mixer such that $h_m(.): \mathbb{R}^n \times \phi \rightarrow \mathbb{R}$ and $Q_{mix}(s_t, \mathbf{u_t}) = h_m\left(Q_{a_1}(),..,Q_{a_n}(), h_p(s_t)\right)$.
QMIX architecture is presented in Figure \ref{fig:qmix}.

The monotonicity of $Q_{mix}$ with respect to the individual $Q_a$ functions is satisfied because a neural network comprised of monotonic functions ($h_m$) and strictly positive weights ($h_p$) is monotonic with respect to its inputs ($Q_a$). 
Since we are in partial observability, individual $Q_a$ networks are RNNs made of GRU \citep{Chung2014EmpiricalModeling}.
The optimisation procedure follows the same principles of the DQN algorithm and the loss applied to $Q_{mix}(s_t, \mathbf{u_t})$ is defined in Equation \ref{eq:QMIX_loss}.

\begin{equation}
    \mathcal{L}(\theta) = \mathds{E}_{\langle s_{t},\mathbf{u_{t}},r_{t},s_{t+1} \rangle \sim B}
    \bigg[  
    \big(r_{t} + \gamma \max_{\mathbf{u} \in \mathcal{U}} Q_{mix}(s_{t+1}, \mathbf{u}; \theta')
    - Q_{mix}(s_{t}, \mathbf{u_{t}}; \theta)\big)^{2}
    \bigg] 
    \label{eq:QMIX_loss}
\end{equation}

Since this method trains recurrent neural networks, the replay buffer does not store isolated transitions $\langle s_{t},\mathbf{u_{t}},r_{t},s_{t+1}\rangle$ but instead stores sequences of contiguous transitions.
Individual $Q_a$ networks are copied as well was as the mixer to produce target networks represented by $\theta'$.

In QMIX implementation, as in QVMix, individual $Q_a$ architecture follows the architecture presented in Figure \ref{fig:indivQ} and takes as input the previous action in addition to the observation.
Note that the hidden states of the recurrent neural network are represented by $h_t$ and their objective is to encode the agent's history $\tau_t$.

\subsection{MAVEN}
\begin{figure}[ht]
\begin{subfigure}[t]{.49\textwidth}
    \centering
    \input{tikz/maven}
    \caption{MAVEN modification of the individual $Q_a$ network.}
    \label{fig:maven}
    \end{subfigure}%
    \hfill
    \begin{subfigure}[t]{0.42\textwidth}
    \centering
    \input{tikz/qvmix}
    \caption{Architecture of the $V_{mix}$ network.}
    \label{fig:qvmix}
    \end{subfigure}%
    
\caption{Details of the MAVEN and QVMix architecture.}
\end{figure}
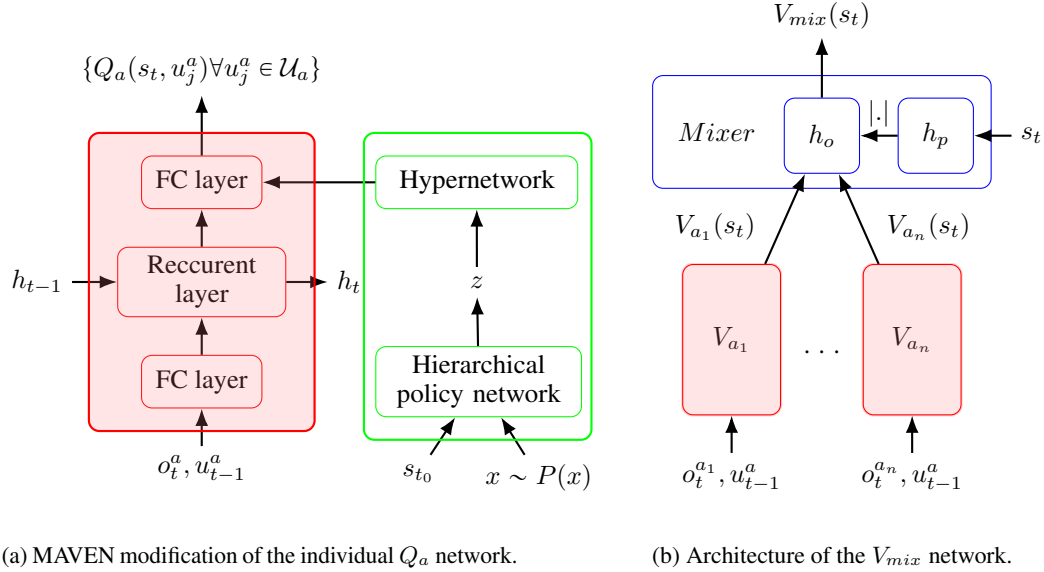

Mahajan et. al. \cite{Mahajan2019MAVEN:Exploration} defined the class of state-joint-action value functions that cannot be represented by QMIX due to its monotonicity constraint.
They demonstrated the existence of payoff matrices in an n-player game with more than three actions per agent for which QMIX learns a suboptimal policy for any training duration, and for epsilon greedy and uniform exploration.
To tackle this problem, in the former QMIX architecture, they added a latent space that conditions individual $Q_a$ function networks with the objective of influencing agent behaviour.
This allows one to learn an ensemble of approximations and therefore an ensemble of policies to improve the exploration capabilities.
The latent variable is the input of a hypernetwork, such as $h_p$ in QMIX, that computes parameters for the fully connected layer linking recurrent cells to outputs in the individual $Q_a$ networks.
This latent variable $z$ is generated per episode and by a hierarchical policy network, taking as input the initial state of the environment together with a random variable (typically discrete and sampled from a uniform distribution).
The latent variable maps the different learnt strategies and the goal of the hierarchical policy network is to select the best strategy based on the initial state $s_0$ which is hypothetically known at testing.
The architecture of the individual $Q_a$ network of MAVEN is represented in Figure \ref{fig:maven}.

MAVEN's network objective function comprises three parts.
To optimise the two hypernetworks (the mixer and the latent space hypernetwork) and the individual recurrent neural networks, a part of the objective is the loss of QMIX defined in Equation \ref{eq:QMIX_loss}.
This loss is computed by fixing the hierarchical policy network and therefore the latent variable $z$.
To optimise the hierarchical policy network, any policy optimisation, such as policy gradient \citep{NIPS1999_464d828b}, computed with the total sum of rewards per episode can be used.
This second objective is computed by fixing both hypernetworks and the individual networks.
To ensure that different values of $z$ imply different behaviours, a mutual information loss between the latent variable and consecutive transitions is added as the third part of the objective.
This third part of the objective requires the introduction of a variational distribution and for further details on the MAVEN optimisation procedure and especially on the construction of the mutual information objective, we refer the reader to \cite{Mahajan2019MAVEN:Exploration}.

\subsection{QVMix}
QVMix \citep{leroy2020qvmix} is an extension of the Deep Quality-Value (DQV) family of algorithms \citep{sabatelli2018deepQV,sabatelli2020deep} to the cooperative MARL setting.
The principle of DQV is to learn both the $Q$ function, $Q(s, u; \theta)$, and the $V$ function, $V(s; \phi)$, simultaneously.

In SARL, following the principles of DQN, both networks are trained with the losses defined in Equations \ref{eq:dqv_v_update} and \ref{eq:dqv_q_update}.

\begin{equation}
    \mathcal{L}(\phi) = \mathds{E}_{\langle 
    s_{t},u_{t},r_{t},s_{t+1}
    \rangle\sim B} 
    \bigg[\big(r_{t} + \gamma V(s_{t+1}; \phi') - V(s_{t}; \phi)\big)^{2}\bigg]
    \label{eq:dqv_v_update}
\end{equation}

\begin{equation}
    \mathcal{L}(\theta) = \mathds{E}_{\langle s_{t},u_{t},r_{t},s_{t+1}
    \rangle\sim B} 
    \bigg[\big(r_{t} + \gamma V(s_{t+1}; \phi') - Q(s_{t}, u_{t}; \theta)\big)^{2}\bigg]
    \label{eq:dqv_q_update}
\end{equation}

In QVMix, the $V$ network is updated with the loss defined in Eq. \ref{eq:dqv_v_update} and has the same architecture as the $Q$ network of QMIX, except that actions are not considered.
The architecture is presented in Figure \ref{fig:qvmix}.
The $Q_{mix}$ network remains the same as in QMIX but is now updated with the loss defined in Eq. \ref{eq:dqv_q_update}.
Again, $B$ stores sequences of contiguous transitions instead of single transitions to train recurrent neural networks.

\section{Elo score}
\label{app:elo}
The purpose of the Elo rating system \citep{elo1978rating} is to assign each player of a population with a rating $R$ to rank them.
From these ratings, one can compute the probability that a player will win when facing another one.
Let $R_A$ and $R_B$ be the ELO scores of player A and B, respectively.
In such a context, the probability that player A (B) wins over player is B (A) is computed using Equation \ref{eq:elo_predict1} (\ref{eq:elo_predict2}) given below.

\begin{align}
    \label{eq:elo_predict1}
    E_A=\frac{10^{R_A/400}}{10^{R_A/400} + 10^{R_B/400}}
    \\
    \label{eq:elo_predict2}
    E_B=\frac{10^{R_B/400}}{10^{R_A/400} + 10^{R_B/400}}
\end{align}%

One can see that $E_A + E_B = 1$.
The number $400$ can be considered as a parameter.
It determines that if the Elo score of player A is 400 points above that of B, it has a ten-times greater chance of defeating B.
In order to update the rating of player $A$ after a game, we take into account its score $S_A$ which is equal to $1$ for a win, $0$ for a loss and $0.5$ for a draw.
The updated score $R'_A$ is defined in Equation \ref{eq:elo_update} where $cst$ is a constant that defines the maximum possible update of the Elo score.
Typically, $cst$ is $32$ but for our experiments, we set it to $10$ to decrease the amplitude of oscillations in the Elo score during tests.

\begin{equation}
    \label{eq:elo_update}
    R'_A = R_A + cst * (S_A - E_A)
\end{equation}

\section{Competitive SMAC}
\label{app:smac}

\begin{figure}[ht]
\begin{subfigure}{.49\textwidth}
    \centering
    \includegraphics[height=68pt]{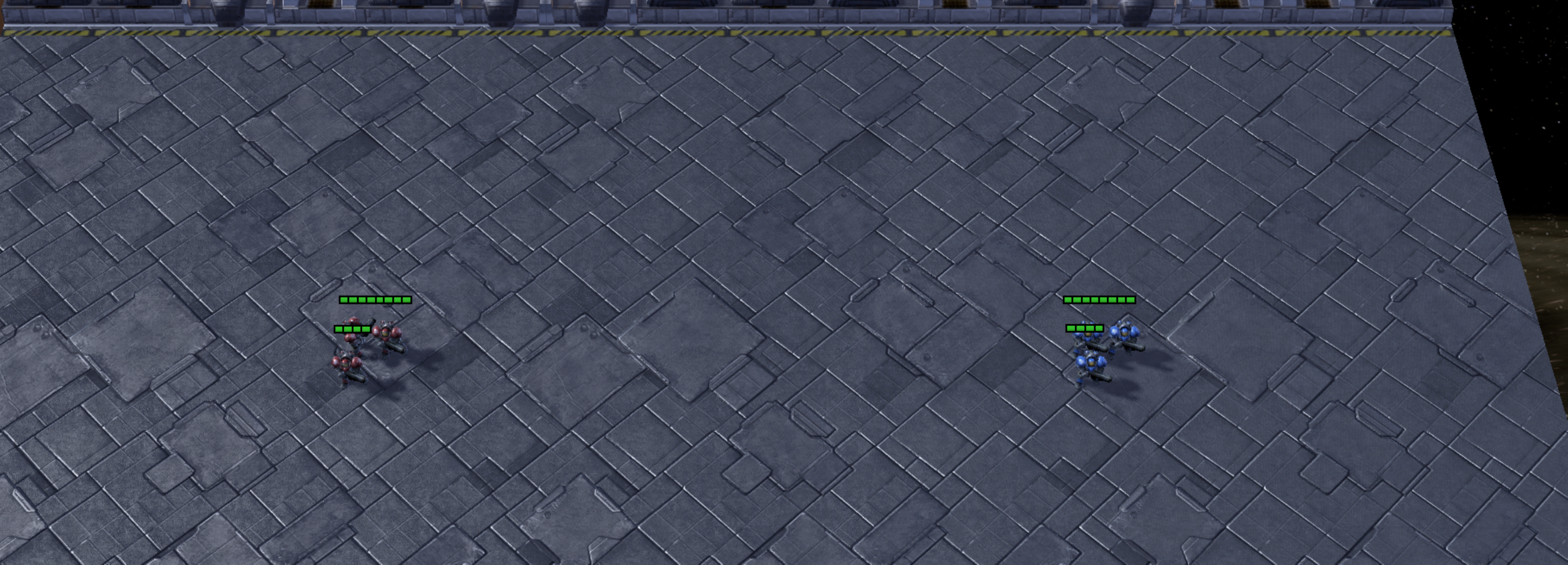}
    \caption{3m}
    \label{fig:3m_map_game}
\end{subfigure}
\begin{subfigure}{.5\textwidth}
        \centering
    \includegraphics[height=68pt]{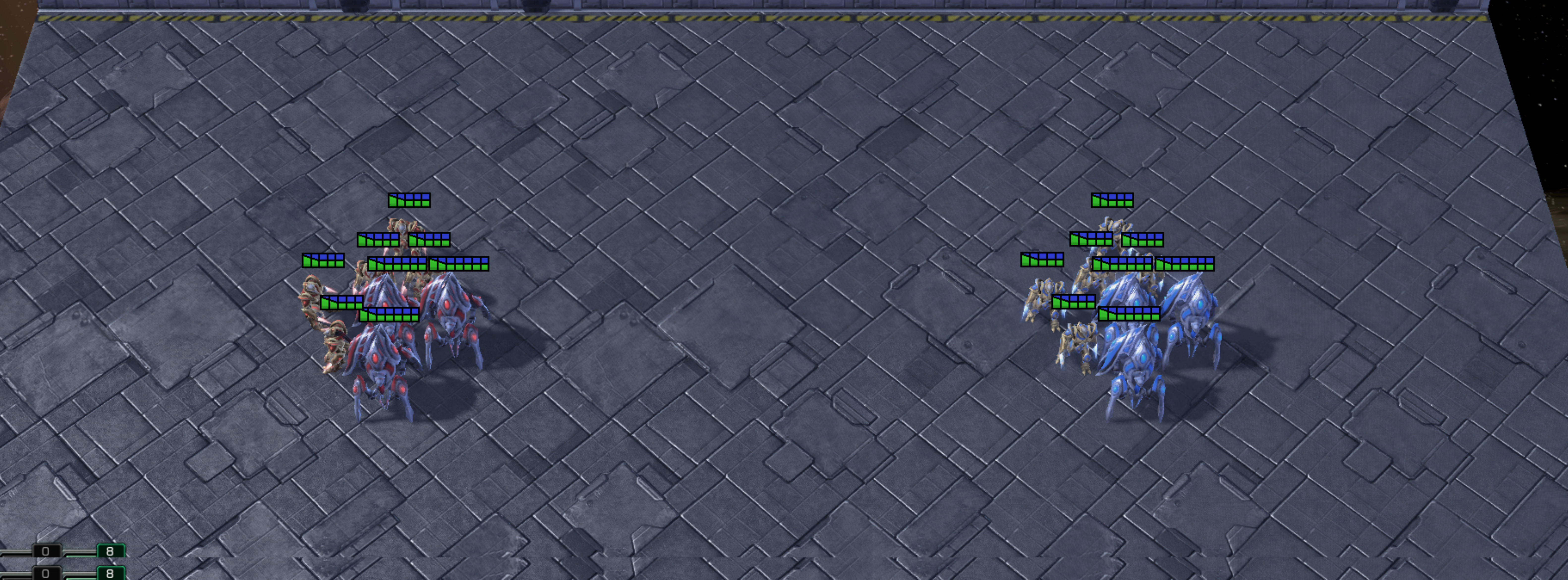}
    \caption{3s5z}
    \label{fig:3s5z_map_game}
\end{subfigure}
\caption{Overview of the 3m and 3s5z SMAC maps seen in StarCraft.}
\label{fig:maps}
\end{figure}

In SMAC, agents have partial observability defined by a sight range, a surrounding circle inside which they can observe allies and enemies and their shooting range is smaller than the sight range.
There are different components in the observation of the state.
An agent observes information about itself: its remaining hit points and shield points, its relative position with respect to the centre of the map and four Booleans representing the direction it can move in (NSWE). 
It also observes information about other agents that are within its sight range: the relative distance, relative x, relative y and the remaining hit points and shield points. 
If the other agent is an ally, it also observes the last action performed by the allied agent. 
If the other agent is an enemy, it observes if the enemy agent is within shooting range.

In the $3m$ map, six marines compete in two teams of three.
A marine has $45$ hit points and shoots at range, inflicting $6$ damage points to an opponent for each attack.
In the $3s5z$ map, six stalkers and ten zealots compete in two teams of eight.
Both units have shield points in addition to hit points.
A shield receives a different amount of damage and regenerates over time if the unit is not attacked again for a given period of time.
A stalker has $80$ hit points and $80$ shield points.
It shoots at range, inflicting $13$ damage points to the shield, $12$ damage points to a zealot's hit points and $17$ to a stalker's hit points.
A zealot has 100 hits points and 50 shield points.
It attacks in melee and inflicts $16$ damage points to the shield and $14$ damage points to the hit points of a zealot or a stalker.
On both maps, there are two possible starting positions for the teams (see Figure \ref{fig:maps}) and agents initially do not see their opponents.

Within both maps, the agent has the choice of eight actions: do nothing, move in one of four directions (NSWE) or attack one of its three opponents.
Some actions are forbidden in SMAC, such as an attack action if the opponent is not within shooting range.
Therefore, before choosing an action, agents must consider which ones are available.

At each timestep, the agent receives a zero or positive reward, common to each agent of the same team. 
This reward is the sum of a zero or positive reward for the damage dealt, a positive reward if an enemy unit's hit points reach zero, and a positive reward if all enemy units are defeated. 
Maximising the reward forces the team to neutralise every unit of the opposing team.

\section{Training parameters}
\label{app:train_param}
Learning parameters of the three methods were determined by default configurations provided by their different authors.
They are the same as the ones used in QVMix implementation \citep{leroy2020qvmix}.
We hereafter provide a description of some of these training parameters.

Individual networks are 64 cells GRU enclosed with fully connected layers (see Fig \ref{fig:indivQ}).
The mixer network is the same as in \citep{Rashid2018} with an embedded size of 32.
The individual and mixer networks are the same for the three methods.
We used the default parameters of MAVEN policy networks provided by \citep{Mahajan2019MAVEN:Exploration}.
For QVMix, the $V$ network is a copy of the QMIX network with only one output for each $V$ network.

For each learning scenario, networks are updated regardless of how episodes have been generated.
Networks are updated from a replay buffer that collects the $5000$ latest played episodes and $32$ of them are sampled from it to update the network.
The network update is performed every eight episodes in the $3m$ map and every episode in the $3s5z$ map.
The difference is justified by the desire to increase the number of network updates for $3s5$ to improve performances, especially against the heuristic.
The epsilon greedy exploration starts with an epsilon equal to $1$ decreasing linearly to $0.05$ during $2$ million timesteps.
This is perhaps the main difference with respect to the provided parameters that decreases the epsilon only during $0.5$ million timesteps.
The discount factor is $\gamma = 0.99$ and the learning rate is $0.0005$.
Target networks are updated every $200$ episodes.
We refer the reader to \citep{Mahajan2019MAVEN:Exploration} for further parameter definitions for MAVEN optimisation.

\section{Training time}
\label{app:train_time}

\begin{figure}[ht]
    \centering
    \includegraphics[width=\textwidth]{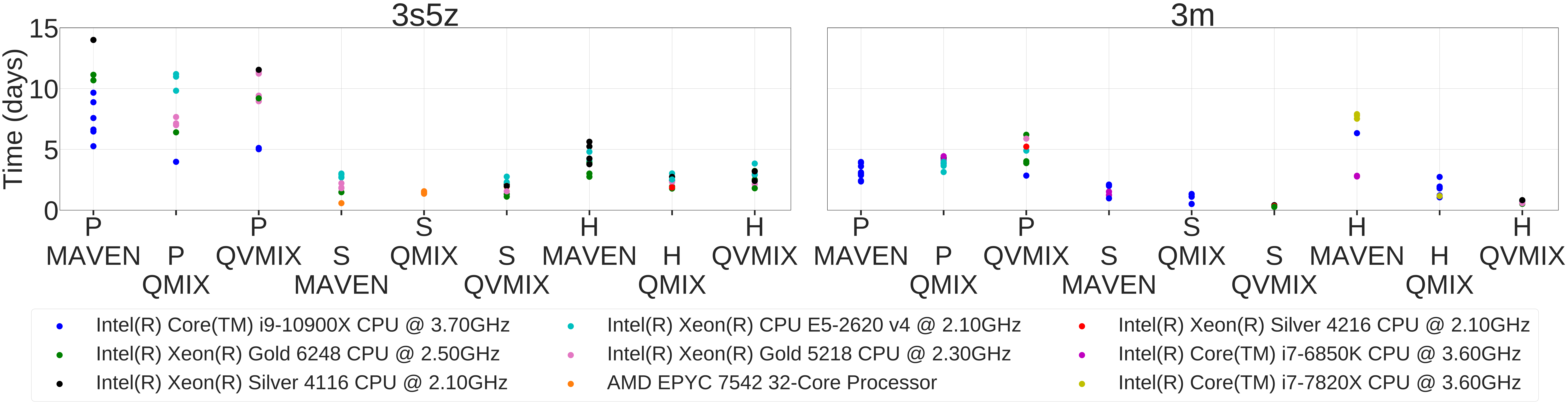}
    \caption{Training duration, in days, for the different learning scenarios tested in this paper.
    The different colours represent the different CPUs that have been used to perform the experiments.}
    \label{fig:training_time}
\end{figure}

Experiments were performed with CPUs only because small recurrent neural networks do not arguably benefit from GPU.
We had access to several types of computer, with different numbers of CPUs accessible at the same time.
Training times for each experiment performed are presented in Figure \ref{fig:training_time}.
With all these different hardware configurations, it is not possible to rigorously compare the times of the experiments.
However, it is possible to present the time complexity.
As explained in Section \ref{sec:experiments}, training in self-play requires five-times fewer environment timesteps than training within a population, but also five-times fewer network updates.
Furthermore, when training a population, networks are updated sequentially, which also increases the time.
Finally, SC2 processes are prone to bugs and therefore sometimes need to be restarted.
As the actions of all agents in the different running environments are performed simultaneously, these restarts are time-consuming operations as the processes have to wait for the faulty one.

\newpage

\section{Additional Elo score boxplots}
\label{app:all_no_h}

\begin{figure*}[ht]
\centering

\begin{subfigure}{\textwidth}
\centering
\includegraphics[width=\textwidth]{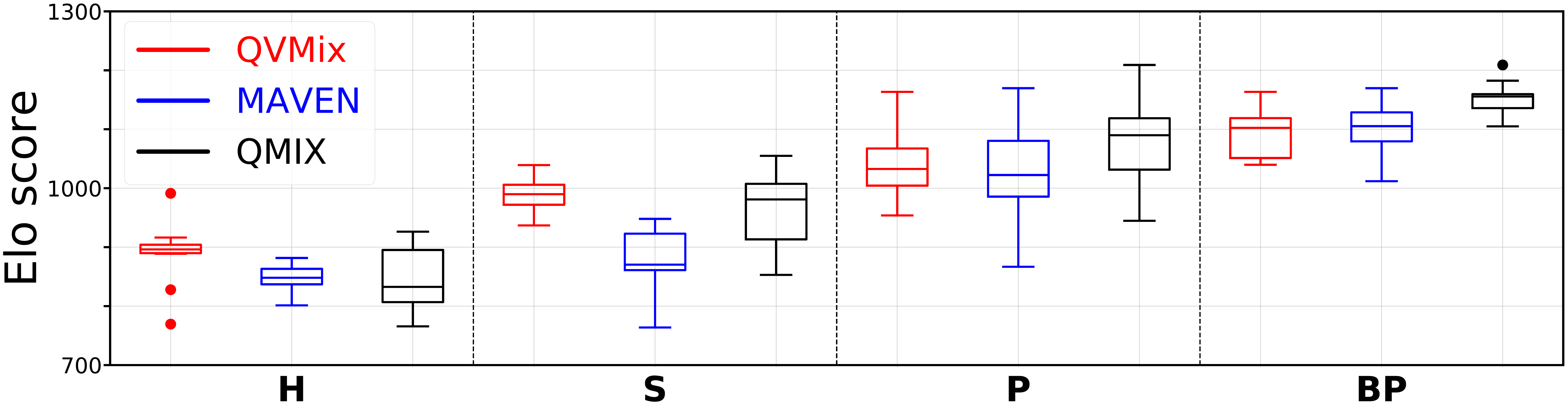}
\caption{$3m$}
\label{subfig:3m_all_no_h}
\end{subfigure}
\begin{subfigure}{\textwidth}
\centering
\includegraphics[width=\textwidth]{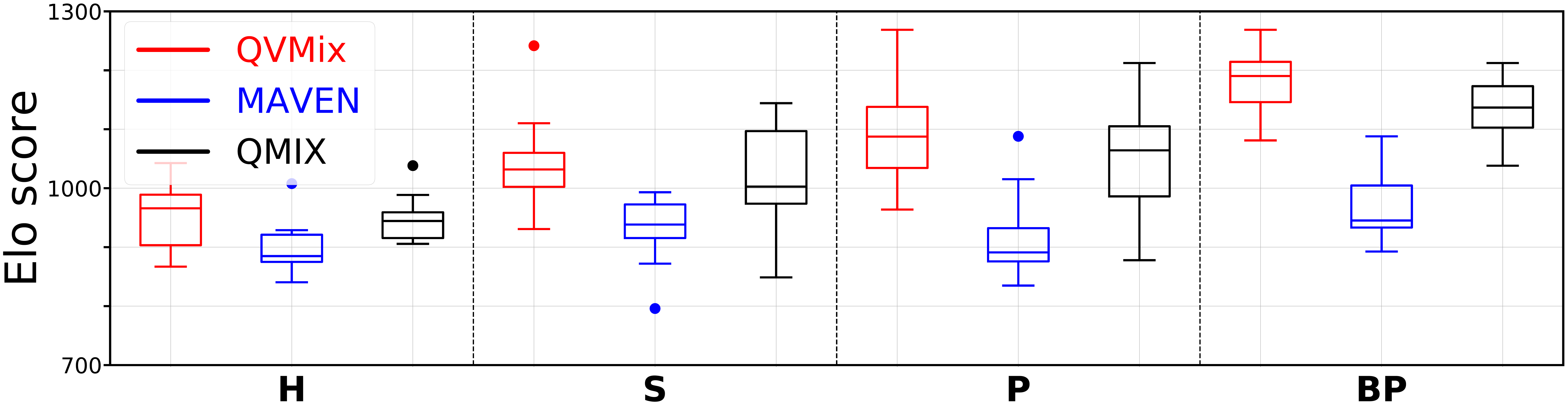}
\caption{$3s5z$}
\label{subfig:3s5z_all_no_h}
\end{subfigure}
\caption{
Elo score box plots of two test populations, in $3m$ at the top and in $3s5z$ at the bottom, composed of teams trained using three methods and three learning scenarios.
The training method is either QVMix (red), MAVEN (blue) or QMIX (black).
Box plots represent the distribution of the Elo scores of teams trained either against the heuristic (\textbf{H}), in self-play (\textbf{S}), within a population (\textbf{P}) or the best of each population (\textbf{BP}).
Box plots present the median, the first quantile ($Q1$) and the third quantile ($Q3$). The reach of whiskers is defined by $1.7*(Q3-Q1)$.
}
\label{fig:all_no_h}
\end{figure*}

Boxplots of the test populations composed of all methods without the heuristic for both maps are presented in Figure \ref{fig:all_no_h}.

\end{document}

%% file: fig_elo.tex
\captionsetup[subfigure]{justification=centering}
    \includegraphics[width=.95\textwidth]{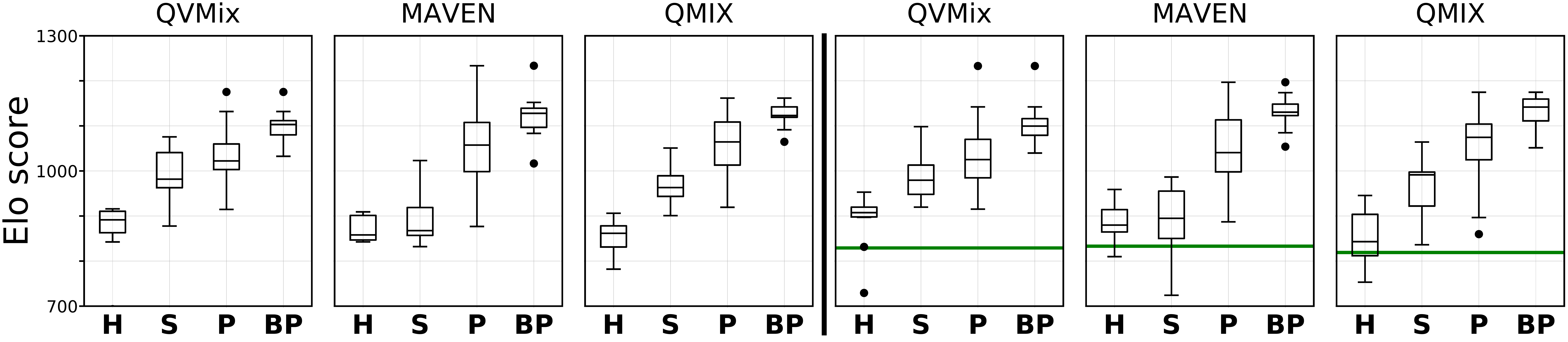}
    \begin{subfigure}{.045\textwidth}
    \centering
    \caption*{}
    \end{subfigure}%
    \begin{subfigure}{.455\textwidth}
        \begin{subfigure}{.33\textwidth}
            \renewcommand\thesubfigure{\alph{subfigure}.1}
          \centering
          \caption{}
          \label{subfig:elo_no_h_methodQVMIX}
        \end{subfigure}%
        \begin{subfigure}{.33\textwidth}
        \addtocounter{subfigure}{-1}
            \renewcommand\thesubfigure{\alph{subfigure}.2}
          \centering
          \caption{}
          \label{subfig:elo_no_h_methodMAVEN}
        \end{subfigure}%
        \begin{subfigure}{.33\textwidth}
        \addtocounter{subfigure}{-1}
            \renewcommand\thesubfigure{\alph{subfigure}.3}
          \centering
          \caption{}
          \label{subfig:elo_no_h_methodQMIX}
        \end{subfigure}%
    \centering
    \addtocounter{subfigure}{-1}
    \caption{$3m$ map \textbf{without} heuristic.}
    \label{subfig:elo_no_h_3m}
    \end{subfigure}%
    \begin{subfigure}{.455\textwidth}
        \begin{subfigure}{.33\textwidth}
        \renewcommand\thesubfigure{\alph{subfigure}.1}
          \centering
          \caption{ }
          \label{subfig:elo_h_methodQVMIX}
        \end{subfigure}%
        \begin{subfigure}{.33\textwidth}
         \addtocounter{subfigure}{-1}
        \renewcommand\thesubfigure{\alph{subfigure}.2}
          \centering
          \caption{ }
          \label{subfig:elo_h_methodMAVEN}
        \end{subfigure}%
        \begin{subfigure}{.33\textwidth}
            \addtocounter{subfigure}{-1}
            \renewcommand\thesubfigure{\alph{subfigure}.3}
          \centering
          \caption{ }
          \label{subfig:elo_h_methodQMIX}
        \end{subfigure}%
    \centering
    \addtocounter{subfigure}{-1}
    \caption{$3m$ map \textbf{with} heuristic.}
     \label{subfig:elo_h_3m}
    \end{subfigure}%

    \includegraphics[width=.95\textwidth]{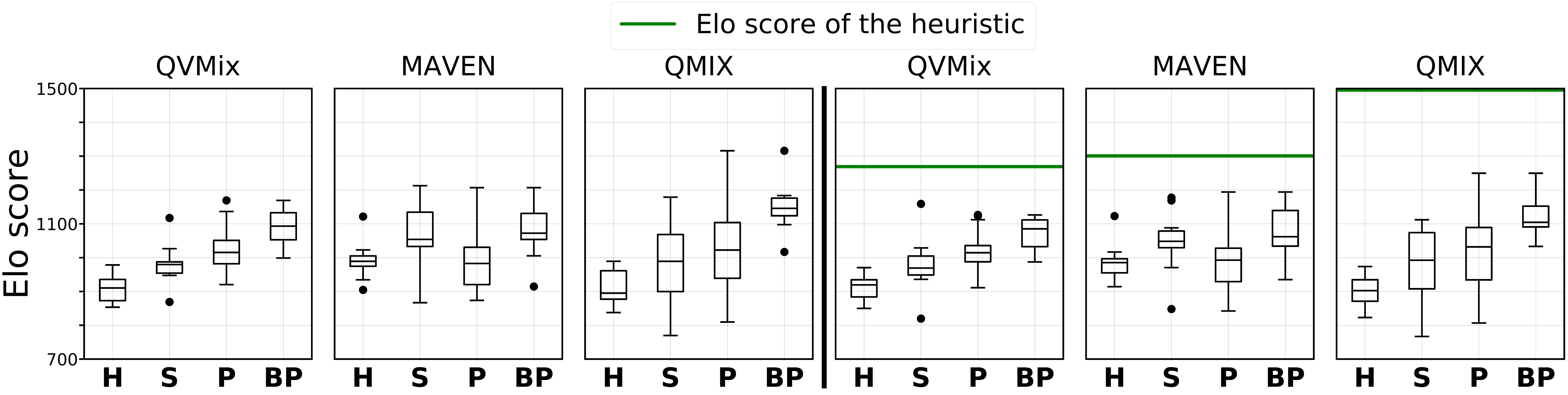}
    \begin{subfigure}{.045\textwidth}
    \centering
    \caption*{}
    \end{subfigure}%
    \begin{subfigure}{.455\textwidth}
        \begin{subfigure}{.33\textwidth}
            \renewcommand\thesubfigure{\alph{subfigure}.1}
          \centering
          \caption{}
          \label{subfig:3s5z_elo_no_h_methodQVMIX}
        \end{subfigure}%
        \begin{subfigure}{.33\textwidth}
        \addtocounter{subfigure}{-1}
            \renewcommand\thesubfigure{\alph{subfigure}.2}
          \centering
          \caption{}
          \label{subfig:3s5z_elo_no_h_methodMAVEN}
        \end{subfigure}%
        \begin{subfigure}{.33\textwidth}
        \addtocounter{subfigure}{-1}
            \renewcommand\thesubfigure{\alph{subfigure}.3}
          \centering
          \caption{}
          \label{subfig:3s5z_elo_no_h_methodQMIX}
        \end{subfigure}%
    \centering
    \addtocounter{subfigure}{-1}
    \caption{$3s5z$ map \textbf{without} heuristic.}
     \label{subfig:elo_no_h_3s5z}
    \end{subfigure}%
    \begin{subfigure}{.455\textwidth}
        \begin{subfigure}{.33\textwidth}
        \renewcommand\thesubfigure{\alph{subfigure}.1}
          \centering
          \caption{ }
          \label{subfig:3s5z_elo_h_methodQVMIX}
        \end{subfigure}%
        \begin{subfigure}{.33\textwidth}
         \addtocounter{subfigure}{-1}
        \renewcommand\thesubfigure{\alph{subfigure}.2}
          \centering
          \caption{ }
          \label{subfig:3s5z_elo_h_methodMAVEN}
        \end{subfigure}%
        \begin{subfigure}{.33\textwidth}
            \addtocounter{subfigure}{-1}
            \renewcommand\thesubfigure{\alph{subfigure}.3}
          \centering
          \caption{ }
          \label{subfig:3s5z_elo_h_methodQMIX}
        \end{subfigure}%
    \centering
    \addtocounter{subfigure}{-1}
    \caption{$3s5z$ map \textbf{with} heuristic.}
     \label{subfig:elo_h_3s5z}
    \end{subfigure}%
    
    \caption{Elo score box plots of 12 test populations. 
    Half of the experiments were performed in the $3m$ map, shown at the top (\ref{subfig:elo_no_h_3m}, \ref{subfig:elo_h_3m}) and the other half in the $3s5z$ map, shown at the bottom (\ref{subfig:elo_no_h_3s5z}, \ref{subfig:elo_no_h_3s5z}). 
    In each test population, teams are trained with the same method, which is either QVMix (\ref{subfig:elo_no_h_methodQVMIX}, \ref{subfig:elo_h_methodQVMIX},\ref{subfig:3s5z_elo_no_h_methodQVMIX}, \ref{subfig:3s5z_elo_h_methodQVMIX}), MAVEN (\ref{subfig:elo_no_h_methodMAVEN}, \ref{subfig:elo_h_methodMAVEN},\ref{subfig:3s5z_elo_no_h_methodMAVEN}, \ref{subfig:3s5z_elo_h_methodMAVEN}) or QMIX (\ref{subfig:elo_no_h_methodQMIX}, \ref{subfig:elo_h_methodQMIX},\ref{subfig:3s5z_elo_no_h_methodQMIX}, \ref{subfig:3s5z_elo_h_methodQMIX}).
    In \ref{subfig:elo_h_3m} and \ref{subfig:elo_h_3s5z}, the heuristic is present in the test population and a green line represents its Elo score.
    Box plots represent the distribution of the ELO scores of the teams trained either against the heuristic (\textbf{H}), in self-play (\textbf{S}), within a population (\textbf{P}) or the best of each training population (\textbf{BP}).
    For most methods, teams trained within a population achieved the highest Elo scores.
    Box plots present the median, the first quantile ($Q1$) and the third quantile ($Q3$). The reach of whiskers is defined by $1.7*(Q3-Q1)$.
    }
    \label{fig:elo_method}

%% file: fig_training.tex
\begin{subfigure}{\textwidth}
\includegraphics[width=\textwidth]{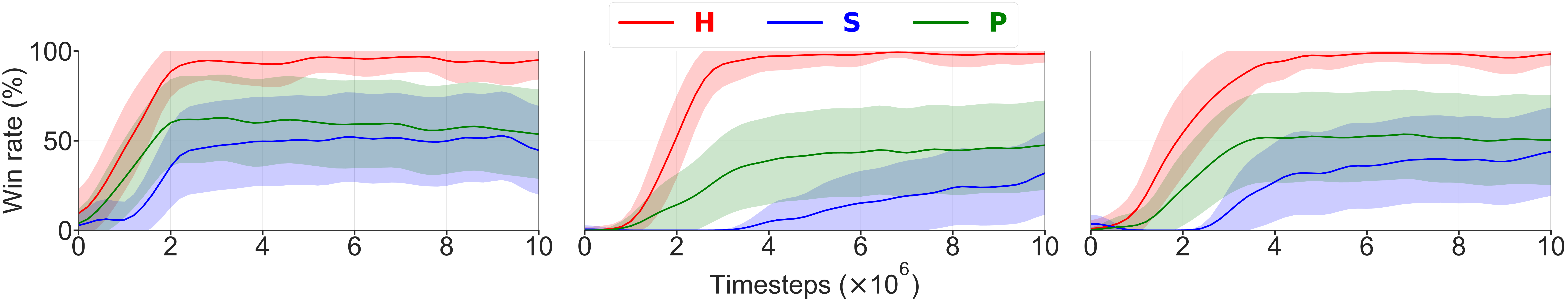}
    \begin{subfigure}{.05\textwidth}
    \centering
    \caption*{}
    \end{subfigure}%
    \begin{subfigure}{.31\textwidth}
    \renewcommand\thesubfigure{\alph{subfigure}.1}
      \centering
      \caption{QVMix}
      \label{subfig:vs_h_methodQVMIX}
    \end{subfigure}%
    \begin{subfigure}{.31\textwidth}
    \addtocounter{subfigure}{-1}
    \renewcommand\thesubfigure{\alph{subfigure}.2}
      \centering
      \caption{MAVEN}
      \label{subfig:vs_h_methodMAVEN}
    \end{subfigure}%
    \begin{subfigure}{.31\textwidth}
    \addtocounter{subfigure}{-1}
    \renewcommand\thesubfigure{\alph{subfigure}.3}
      \centering
      \caption{QMIX}
      \label{subfig:vs_h_methodQMIX}
    \end{subfigure}
\addtocounter{subfigure}{-1}
\caption{Win rates achieved against the heuristic in the $3m$ map.}    
\label{subfig:3m_vs_h}
\end{subfigure}
\begin{subfigure}{\textwidth}
    
    \includegraphics[width=\textwidth]{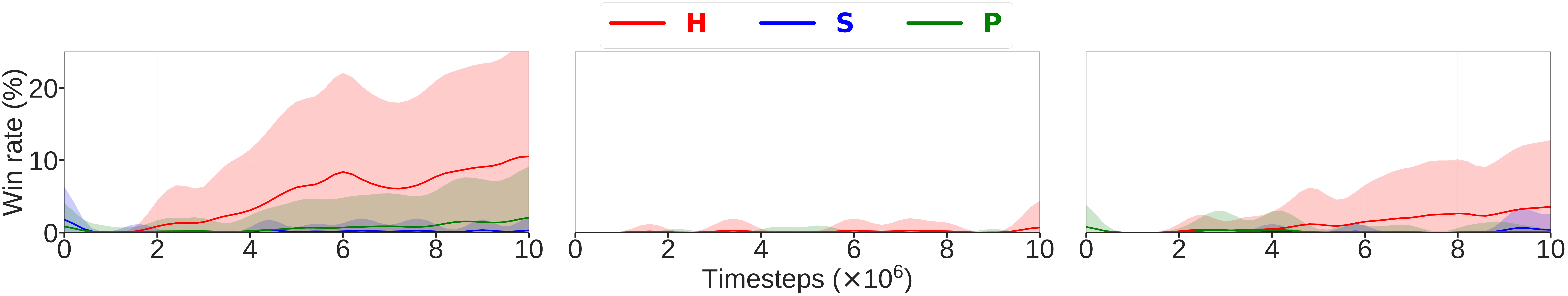}
    \begin{subfigure}{.05\textwidth}
    \centering
    \caption*{}
    \end{subfigure}%
    \begin{subfigure}{.31\textwidth}
    \renewcommand\thesubfigure{\alph{subfigure}.1}
      \centering
      \caption{QVMix}
      \label{subfig:3s5z_vs_h_methodQVMIX}
    \end{subfigure}%
    \begin{subfigure}{.31\textwidth}
    \addtocounter{subfigure}{-1}
    \renewcommand\thesubfigure{\alph{subfigure}.2}
      \centering
      \caption{MAVEN}
      \label{subfig:3s5z_vs_h_methodMAVEN}
    \end{subfigure}%
    \begin{subfigure}{.31\textwidth}
    \addtocounter{subfigure}{-1}
    \renewcommand\thesubfigure{\alph{subfigure}.3}
      \centering
      \caption{QMIX}
      \label{subfig:3s5z_vs_h_methodQMIX}
    \end{subfigure}
\addtocounter{subfigure}{-1}
\caption{Win rates achieved against the heuristic in the $3s5z$ map. Note the change in scale of the y-axis.}
\label{subfig:3s5z_vsh}
\end{subfigure}
\begin{subfigure}{\textwidth}
    \includegraphics[width=\textwidth]{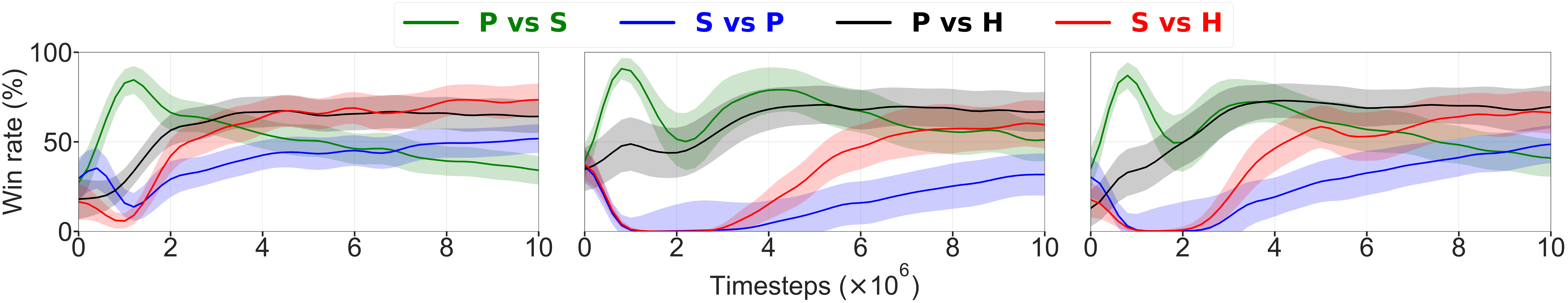}
    \begin{subfigure}{.05\textwidth}
    \centering
    \caption*{}
    \end{subfigure}%
    \begin{subfigure}{.31\textwidth}
    \renewcommand\thesubfigure{\alph{subfigure}.1}
      \centering
      \caption{QVMix}
      \label{subfig:duo_methodQVMIX}
    \end{subfigure}%
    \begin{subfigure}{.31\textwidth}
    \addtocounter{subfigure}{-1}
    \renewcommand\thesubfigure{\alph{subfigure}.2}
      \centering
      \caption{MAVEN}
      \label{subfig:duo_methodMAVEN}
    \end{subfigure}%
    \begin{subfigure}{.31\textwidth}
    \addtocounter{subfigure}{-1}
    \renewcommand\thesubfigure{\alph{subfigure}.3}
      \centering
      \caption{QMIX}
      \label{subfig:duo_methodQMIX}
    \end{subfigure}
\addtocounter{subfigure}{-1}
\caption{Win rates achieved by teams trained in the $3m$ map against themselves.}
\label{subfig:3m_duo}
\end{subfigure}
\begin{subfigure}{\textwidth}
    \includegraphics[width=\textwidth]{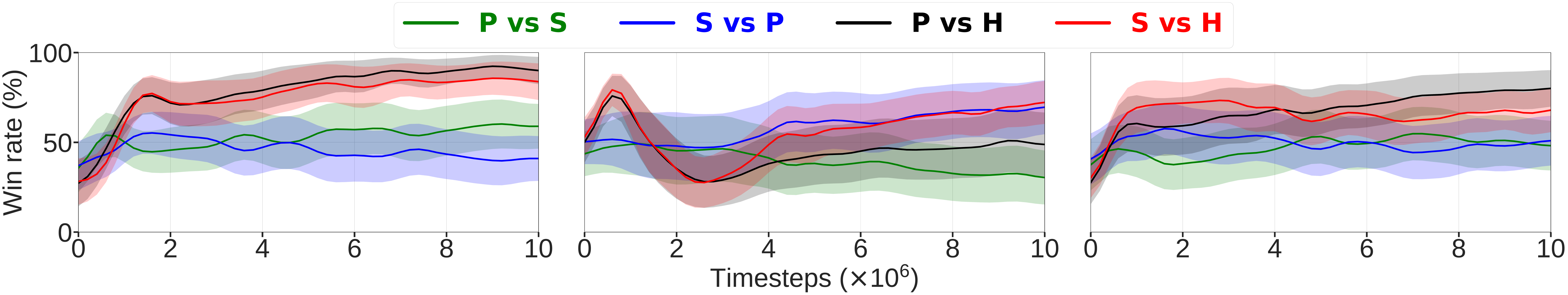}
    \begin{subfigure}{.05\textwidth}
    \centering
    \caption*{}
    \end{subfigure}%
    \begin{subfigure}{.31\textwidth}
    \renewcommand\thesubfigure{\alph{subfigure}.1}
      \centering
      \caption{QVMix}
      \label{subfig:3s5z_duo_methodQVMIX}
    \end{subfigure}%
    \begin{subfigure}{.31\textwidth}
    \addtocounter{subfigure}{-1}
    \renewcommand\thesubfigure{\alph{subfigure}.2}
      \centering
      \caption{MAVEN}
      \label{subfig:3s5z_duo_methodMAVEN}
    \end{subfigure}%
    \begin{subfigure}{.31\textwidth}
    \addtocounter{subfigure}{-1}
    \renewcommand\thesubfigure{\alph{subfigure}.3}
      \centering
      \caption{QMIX}
      \label{subfig:3s5z_duo_methodQMIX}
    \end{subfigure}
\addtocounter{subfigure}{-1}
\caption{Win rates achieved by teams trained in the $3s5z$ map against themselves.}
\label{subfig:3s5z_duo}
\end{subfigure}

\caption{
Means of win rate achieved along training timesteps by confronting teams trained with the same method against the heuristic (\ref{subfig:3m_vs_h},\ref{subfig:3s5z_vsh}) or against other teams trained with a different learning scenario (\ref{subfig:3m_duo}, \ref{subfig:3s5z_duo}).
Teams are trained either with QVMix (\ref{subfig:vs_h_methodQVMIX}, \ref{subfig:duo_methodQVMIX},\ref{subfig:3s5z_vs_h_methodQVMIX}, \ref{subfig:3s5z_duo_methodQVMIX}), MAVEN (\ref{subfig:vs_h_methodMAVEN}, \ref{subfig:duo_methodMAVEN},\ref{subfig:3s5z_vs_h_methodMAVEN}, \ref{subfig:3s5z_duo_methodMAVEN}) or QMIX (\ref{subfig:vs_h_methodQMIX}, \ref{subfig:duo_methodQMIX},\ref{subfig:3s5z_vs_h_methodQMIX}, \ref{subfig:3s5z_duo_methodQMIX}).
Tests were performed in the $3m$ map, shown at the top, and in the $3s5z$ maps, shown at the bottom. 
In \ref{subfig:3m_vs_h} and \ref{subfig:3s5z_vsh}, win rates against the heuristic are presented in red, blue and green for teams trained against the heuristic, in self-play and within a population, respectively.
In \ref{subfig:3m_duo} and \ref{subfig:3s5z_duo}, win rates of teams trained within a population against teams trained in self-play are presented in green and against teams trained against the heuristic in black.
Win rates of teams trained in self-play against teams trained within a population are presented in blue and against teams trained against the heuristic in red.
Training against the heuristic achieves the best win rates against the heuristic with the lowest variance, but is the worst scenario when competing against other trained teams.
The error band is half the standard deviation.
} 
\label{fig:heuristic_plot}

%% file: tikz/qmix.tex
\begin{tikzpicture}[node distance=.5cm]

\tikzstyle{netbox} = [rectangle, rounded corners, minimum width=1cm, minimum height=1cm, draw=blue]
\tikzstyle{mixerbox} = [rectangle, rounded corners, minimum width=4.5cm, minimum height=1.5cm, draw=blue]
\tikzstyle{indixQbox} = [rectangle, rounded corners, minimum width=1.3cm, minimum height=2cm, draw=red]
\tikzstyle{io} = [minimum width=0.2cm,minimum height=0.5cm]
\tikzstyle{emptynetbox} = [rectangle, rounded corners, minimum width=1cm, minimum height=1cm]

\node (output_node) [io] {$Q_{mix}(s_t, \mathbf{u_t})$};
\node (mixerbox) [mixerbox, below=of output_node, yshift=-0cm, label={[xshift=-1.4cm, yshift=-1cm]$Mixer$}] {};
\node (indivQbox1) [indixQbox, below=of mixerbox, xshift=-1.2cm, yshift=-.5cm, label={[xshift=-0.22cm, yshift=.2cm]$Q_{a_1}(s_t, u^1_t)$}] {$Q_{a_1}$};
\node (indivQboxn) [indixQbox, below=of mixerbox, xshift=1.2cm, yshift=-.5cm, label={[xshift=0.22cm, yshift=.2cm]$Q_{a_n}(s_t, u^n_t)$}] {$Q_{a_n}$};
\node (mixer_net) [netbox, below=of output_node, yshift=-0.3cm] {$h_m$};
\node (param_net) [netbox, right=of mixer_net] {$h_p$};
\node (param_net2) [emptynetbox, left=of mixer_net] {};

\node (intput_node1) [io,below=of indivQbox1, yshift=-0cm, label={[xshift=1.2cm, yshift=1.2cm]. . .}] {$o^{a_1}_t, u^a_{t-1}$};
\node (intput_noden) [io,below=of indivQboxn, yshift=-0cm] {$o^{a_n}_t, u^a_{t-1}$};
\node (intput_nodestate) [io, right=of param_net] {$s_t$};
\node (intput_nodestate2) [io, left=of param_net2] {};

\draw [-Latex, thick] (intput_node1) -- (indivQbox1);
\draw [-Latex, thick] (intput_noden) -- (indivQboxn);
\draw [-Latex, thick] (indivQbox1) --  (mixer_net);
\draw [-Latex, thick] (indivQboxn) -- (mixer_net);
\draw [-Latex, thick] (param_net) -- (mixer_net) node[midway, yshift=0.3cm] {$|.|$};
\draw [-Latex, thick] (intput_nodestate) -- (param_net);
\draw [-Latex, thick] (mixer_net) -- (output_node);

\draw[thick, rounded corners, draw=red, fill=red!50, opacity=0.2] ($(indivQbox1.north west)$) rectangle ($(indivQbox1.south east)$);
\draw[thick, rounded corners, draw=red, fill=red!50, opacity=0.2] ($(indivQboxn.north west)$) rectangle ($(indivQboxn.south east)$);
\end{tikzpicture}

%% file: tikz/indivQ.tex
\begin{tikzpicture}[node distance=.5cm]

\tikzstyle{netbox} = [rectangle, rounded corners, minimum width=2cm, minimum height=0.5cm,text centered, draw=red]
\tikzstyle{io} = [minimum width=0.2cm,minimum height=0.5cm, text centered]

\node (output_node) [io] {$\{Q_a(s_t, u^a_{j}) \forall u^a_j \in \mathcal{U}_a \}$};
\node (fctop) [netbox, below=of output_node, yshift=-0.3cm] {FC layer};
\node (rec_net) [netbox, below=of fctop, text width=2cm] {Reccurent layer};
\node (fcbot) [netbox, below=of rec_net] {FC layer};
\node (input_node) [io,below=of fcbot, yshift=-0.3cm] {$o^a_t, u^a_{t-1}$};
\node (input_rec) [io,left=of rec_net, xshift=-0.3cm] {$h_{t-1}$};
\node (output_rec) [io,right=of rec_net, xshift=+0.3cm] {$h_t$};

\draw [-Latex, thick] (input_node) -- (fcbot);
\draw [-Latex, thick] (fcbot) -- (rec_net);
\draw [-Latex, thick] (rec_net) -- (fctop);
\draw [-Latex, thick] (fctop) -- (output_node);
\draw [-Latex, thick] (input_rec) -- (rec_net);
\draw [-Latex, thick] (rec_net) -- (output_rec);

\draw[thick, rounded corners, draw=red, fill=red!50, opacity=0.2] ($(fctop.north west)+(-0.5,0.5)$) rectangle ($(fcbot.south east)+(0.5,-0.5)$);
\draw[thick, rounded corners, draw=red] ($(fctop.north west)+(-0.5,0.5)$) rectangle ($(fcbot.south east)+(0.5,-0.5)$);
\end{tikzpicture}

%% file: tikz/maven.tex
\begin{tikzpicture}[node distance=.5cm]

\tikzstyle{netbox} = [rectangle, rounded corners, minimum width=1.6cm, minimum height=0.7cm,text centered, draw=red]
\tikzstyle{netboxMaven} = [rectangle, rounded corners, minimum width=2cm, minimum height=0.7cm,text centered, draw=green]
\tikzstyle{io} = [text centered]

\node (output_node) [io] {$\{Q_a(s_t, u^a_{j}) \forall u^a_j \in \mathcal{U}_a \}$};
\node (fctop) [netbox, below=of output_node, yshift=-0.3cm] {FC layer};
\node (rec_net) [netbox, below=of fctop, text width=2cm] {Reccurent layer};
\node (fcbot) [netbox, below=of rec_net] {FC layer};
\node (intput_node) [io,below=of fcbot, yshift=-0cm] {$o^a_t, u^a_{t-1}$};
\node (input_rec) [io,left=of rec_net, xshift=-0.1cm] {$h_{t-1}$};
\node (output_rec) [io,right=of rec_net, xshift=+0.05cm] {$h_t$};

\draw [-Latex, thick] (intput_node) -- (fcbot);
\draw [-Latex, thick] (fcbot) -- (rec_net);
\draw [-Latex, thick] (rec_net) -- (fctop);
\draw [-Latex, thick] (fctop) -- (output_node);
\draw [-Latex, thick] (input_rec) -- (rec_net);
\draw [-Latex, thick] (rec_net) -- (output_rec);

\node (param_net) [netboxMaven, right=of fctop, xshift=1cm, text width=2.45cm] {Hypernetwork};
\node (latentvar) [io, right=of rec_net, xshift=.66cm, text width=2.5cm] {$z$};
\node (latent_net) [netboxMaven, right=of fcbot, xshift=1cm, text width=2.5cm] {Hierarchical policy network};
\node (intput_latent_1) [io,below=of latent_net, yshift=-0cm, xshift=-0.8cm] {$s_{t_0}$};
\node (intput_latent_2) [io,below=of latent_net, yshift=-0cm, xshift=0.8cm] {$x \sim P(x)$};
\draw [-Latex, thick] (param_net) -- (fctop);
\draw [-Latex, thick] (latentvar) -- (param_net);
\draw [-Latex, thick] (latent_net) -- (latentvar);
\draw [-Latex, thick] (intput_latent_1) -- (latent_net);
\draw [-Latex, thick] (intput_latent_2) -- (latent_net);

\draw[thick, rounded corners, draw=red, fill=red!50, opacity=0.2] ($(fctop.north)+(-1.5,0.3)$) rectangle ($(fcbot.south)+(1.5,-0.3)$);
\draw[thick, rounded corners, draw=red] ($(fctop.north)+(-1.5,0.3)$) rectangle ($(fcbot.south)+(1.5,-0.3)$);
\draw[thick, rounded corners, draw=green] ($(param_net.north)+(-1.5,0.3)$) rectangle ($(latent_net.south)+(1.5,-0.3)$);

\end{tikzpicture}

%% file: tikz/qvmix.tex
\begin{tikzpicture}[node distance=.5cm]

\tikzstyle{netbox} = [rectangle, rounded corners, minimum width=1cm, minimum height=1cm, draw=blue]
\tikzstyle{mixerbox} = [rectangle, rounded corners, minimum width=4.5cm, minimum height=1.5cm, draw=blue]
\tikzstyle{indixQbox} = [rectangle, rounded corners, minimum width=1.3cm, minimum height=2cm, draw=red]
\tikzstyle{io} = [minimum width=0.2cm,minimum height=0.5cm]
\tikzstyle{emptynetbox} = [rectangle, rounded corners, minimum width=1cm, minimum height=1cm]

\node (output_node) [io] {$V_{mix}(s_t)$};
\node (mixerbox) [mixerbox, below=of output_node, yshift=-0cm, label={[xshift=-1.4cm, yshift=-1cm]$Mixer$}] {};
\node (indivQbox1) [indixQbox, below=of mixerbox, xshift=-1.2cm, yshift=-.5cm, label={[xshift=-0.22cm, yshift=.2cm]$V_{a_1}(s_t)$}] {$V_{a_1}$};
\node (indivQboxn) [indixQbox, below=of mixerbox, xshift=1.2cm, yshift=-.5cm, label={[xshift=0.22cm, yshift=.2cm]$V_{a_n}(s_t)$}] {$V_{a_n}$};
\node (mixer_net) [netbox, below=of output_node, yshift=-0.3cm] {$h_o$};
\node (param_net) [netbox, right=of mixer_net] {$h_p$};
\node (param_net2) [emptynetbox, left=of mixer_net] {};

\node (intput_node1) [io,below=of indivQbox1, yshift=-0cm, label={[xshift=1.2cm, yshift=1.2cm]. . .}] {$o^{a_1}_t, u^a_{t-1}$};
\node (intput_noden) [io,below=of indivQboxn, yshift=-0cm] {$o^{a_n}_t, u^a_{t-1}$};
\node (intput_nodestate) [io, right=of param_net] {$s_t$};
\node (intput_nodestate2) [io, left=of param_net2] {};

\draw [-Latex, thick] (intput_node1) -- (indivQbox1);
\draw [-Latex, thick] (intput_noden) -- (indivQboxn);
\draw [-Latex, thick] (indivQbox1) --  (mixer_net);
\draw [-Latex, thick] (indivQboxn) -- (mixer_net);
\draw [-Latex, thick] (param_net) -- (mixer_net) node[midway, yshift=0.3cm] {$|.|$};
\draw [-Latex, thick] (intput_nodestate) -- (param_net);
\draw [-Latex, thick] (mixer_net) -- (output_node);

\draw[thick, rounded corners, draw=red, fill=red!50, opacity=0.2] ($(indivQbox1.north west)$) rectangle ($(indivQbox1.south east)$);
\draw[thick, rounded corners, draw=red, fill=red!50, opacity=0.2] ($(indivQboxn.north west)$) rectangle ($(indivQboxn.south east)$);
\end{tikzpicture}